\documentclass[final]{cvpr}

\usepackage{times}
\usepackage{epsfig}
\usepackage{graphicx}
\usepackage{amsmath}
\usepackage{amssymb}

\usepackage{booktabs}
\usepackage{arydshln}
\usepackage{url}
\usepackage{amsfonts}
\usepackage{nicefrac}
\usepackage{microtype}
\usepackage{subcaption}

\usepackage{multirow}
\usepackage{arydshln}
\usepackage{appendix}
\usepackage[dvipsnames]{xcolor}
\usepackage{wrapfig}
\usepackage{float}
\usepackage{adjustbox}
\usepackage[normalem]{ulem}
\usepackage{comment}
\usepackage{flushend}
\usepackage{multicol}
\usepackage{balance}
\usepackage{multibib}
\usepackage{environ}
\newcommand{\acksection}{\section*{Acknowledgments and Disclosure of Funding}}
\NewEnviron{ack}{%
  \acksection
  \BODY
}
\usepackage[pagebackref=true,breaklinks=true,colorlinks,bookmarks=false]{hyperref}
\newcites{Appendix}{Additional References}

\def\cvprPaperID{4174} % *** Enter the CVPR Paper ID here
\def\confYear{CVPR 2021}
\newcommand{\bdata}{\mathcal{C}_{base}}
\newcommand{\ndata}{\mathcal{C}_{novel}}
\newcommand{\classdata}{\mathcal{D}^{class}}

\newcommand{\bdetdata}{\mathcal{D}_{base}}
\newcommand{\ndetdata}{\mathcal{D}_{novel}}

\newcommand{\W}{W}

\DeclareMathOperator*{\argmax}{arg\,max}

\newcommand*{\RG}{\textcolor{ForestGreen}}

\newcommand\blfootnote[1]{%
  \begingroup
  \renewcommand\thefootnote{}\footnote{#1}%
  \addtocounter{footnote}{-1}%
  \endgroup
}
\begin{document}

%%%%%%%%% TITLE
\title{UniT: Unified Knowledge Transfer for Any-shot \\ Object Detection and Segmentation}

\author{
Siddhesh Khandelwal$^{*,1,2}$ \hspace{0.25in} Raghav Goyal$^{*,1,2}$ \hspace{0.25in} Leonid Sigal$^{1,2,3}$\\
%  University of British Columbia\\
% \texttt{\{skhandel, rgoyal14, lsigal\}@cs.ubc.ca}\\
$^1$Department of Computer Science, University of British Columbia\\
Vancouver, BC, Canada \\
$^2$Vector Institute for AI \hspace{1in}   $^3$CIFAR AI Chair \\
\texttt{skhandel@cs.ubc.ca} \hspace{0.25in}
\texttt{rgoyal14@cs.ubc.ca} \hspace{0.25in}
\texttt{lsigal@cs.ubc.ca} \\
}
\maketitle

%%%%%%%%% ABSTRACT
\begin{abstract}
   %  Recent
Methods for object detection and segmentation rely on large scale instance-level annotations for training, which are difficult and time-consuming to collect. Efforts to alleviate this look at varying degrees and quality of supervision.
Weakly-supervised approaches draw on image-level labels to build detectors/segmentors, while zero/few-shot methods assume abundant instance-level data for a set of \emph{base} classes, and none to a few examples for \emph{novel} classes. This taxonomy has largely siloed algorithmic designs. 
In this work, we aim to bridge this divide by proposing an intuitive and unified semi-supervised model that is applicable to a range of supervision: from zero to a few instance-level samples per \emph{novel} class. For \emph{base} classes, our model learns a mapping from weakly-supervised 
to fully-supervised detectors/segmentors.
By learning and leveraging visual and lingual similarities between the \emph{novel} and \emph{base} classes, we transfer 
those mappings to obtain detectors/segmentors for \emph{novel} classes; refining them with a few \emph{novel} class instance-level annotated samples, if available. The overall model is end-to-end trainable and highly flexible. Through extensive experiments on MS-COCO \cite{lin2014microsoft} and Pascal VOC \cite{everingham2010pascal} benchmark datasets we show improved performance in a variety of settings.\blfootnote{$^*$Denotes equal contribution}

\end{abstract}

\vspace{-0.2in}
\section{Introduction}
\vspace{-0.05in}

\begin{figure}[t]
    \centering
    \includegraphics[width=\linewidth]{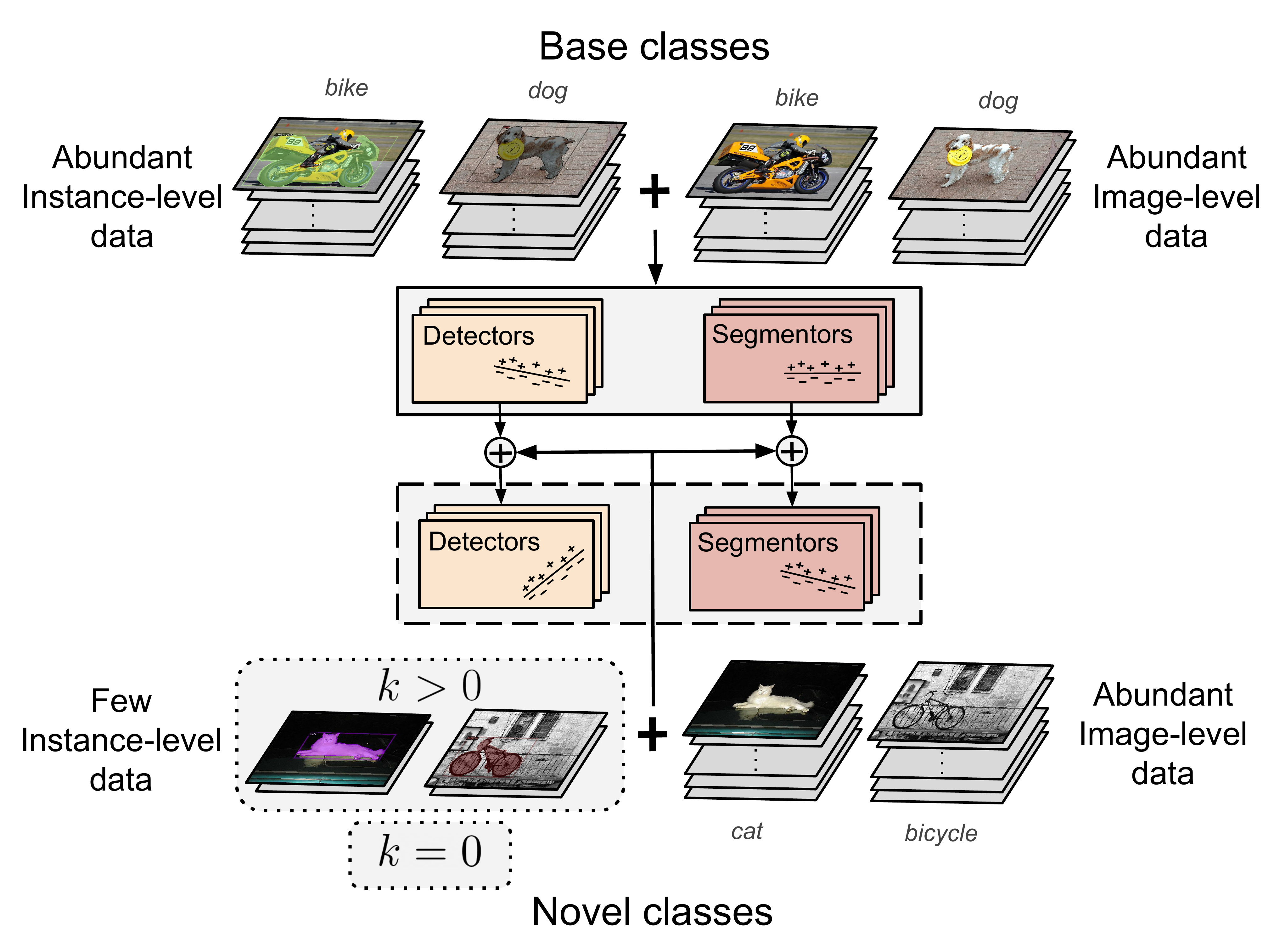}
    \vspace{-0.30in}
    % \caption{Caption is currently commented. Adding it pushes the figure down to 2nd page.}
    \caption{{\bf Semi-supervised Any-shot Detection and Segmentation.} The data used in our setting is categorized in two ways: (1) image-level classification data for \emph{all} the object classes, and (2) abundant instance data for a set of \emph{base} object classes and limited (possibly zero) instance data for a set of \emph{novel} object classes, with the aim to obtain a model that learns to detect/segment both base and novel objects at test time. }
    \label{fig:intro}
    \vspace{-0.25in}
\end{figure}

Over the past decade CNNs % Convolutional Neural Networks (CNNs) 
have emerged as the dominant building blocks for various computer vision understanding tasks, including object classification \cite{he2016deep,russakovsky2015imagenet,sun2017revisiting}, detection \cite{liu2016ssd,redmon2017yolo9000,ren2015faster}, and segmentation \cite{chen2018masklab,he2017mask}. Architectures based on Faster R-CNN \cite{ren2015faster}, Mask R-CNN \cite{he2017mask} and YOLO \cite{redmon2017yolo9000} have achieved impressive performance on a variety of core vision tasks. However, traditional CNN-based approaches %, such as those, 
rely on lots of supervised data for which the annotation efforts can be time-consuming and expensive \cite{hoffman2014lsda, laradji2019masks}. While image-level class labels are easy to obtain, more structured labels such as bounding boxes or segmentations are difficult and expensive\footnote{Segmentation annotations in PASCAL VOC take $239.7$ seconds/image, on average, as compared to $20$ seconds/image for image-level labels \cite{Bearman_2019_ECCV}.}. Further, in certain domains ({\em e.g.}, medical imaging) more detailed labels may require subject expertise. The growing need for efficient learning has motivated development of various approaches and research sub-communities.

On one end of the spectrum, \emph{zero-shot learning} methods require no visual data and use auxiliary information, such as attributes or class names, to form detectors for unseen classes from related seen category detectors %  (learned with abundant data) 
\cite{bansal2018zero, frome2013devise, rahman2018zero, xian2018zero}. \emph{Weakly-supervised learning} methods \cite{arun2019dissimilarity, bilen2016weakly, diba2017weakly, laradji2019masks, wang2018collaborative} aim to utilize readily available coarse image-level labels for more granular downstream tasks, such as object detection \cite{bansal2018zero,rahman2018zero} and segmentation \cite{laradji2019masks,zhou2018weakly}. % \emph{Semi-supervised learning} methods \cite{dong2018few, misra2015watch} explore augmenting small amounts of supervised data with large amounts of unlabelled samples. 
Most recently, \emph{few-shot learning} \cite{andrychowicz2016learning, ravi2016optimization, snell2017prototypical, vinyals2016matching} has emerged as a learning-to-learn paradigm which either learns from few labels directly or by simulation of few-shot learning paradigm through meta-learning \cite{finn2017model,schmidhuber1987evolutionary,thrun2012learning}. % However, most of the existing few-shot learning approaches have addressed efficient learning in coarse image-level classification tasks (with  exception of [cite]); while weakly-supervised methods typically target large scale data applied for learning of more granular tasks, such as object detection and segmentation. 
% However, all of the aforementioned sub-settings have been targeted, to a large extent, by different sub-community of researchers and % classes of algorithms. 
An interesting class of \emph{semi-supervised} methods \cite{gao2019note, hoffman2014lsda, kumar2018dock, tang2016large, uijlings2018revisiting, yang2019detecting} have emerged which aim to % bridge this gap by transferring 
transfer 
knowledge from abundant {\em base} classes to data-starved {\em novel} classes, especially for granular instance-level visual understanding tasks.
However, to date, there isn't a single, unified framework that can effectively leverage various forms and amounts of training data (zero-shot to fully supervised).
%However, to date, there isn't a single, unified framework that can effectively scale to any amount of training data (from zero-shot to fully supervised) 
%[Not sure if this is the right claim]}. \sout{To date, no framework has been developed that can effectively scale to any amount of training data (from zero-shot to fully supervised), especially for granular instance-level visual understanding tasks}. 

We make two fundamental observations that motivate our work. First, image-level supervision is abundant, while instance-level structured labels, such as bounding boxes and segmentation masks, are expensive and scarce. This
is reflected in the scales of widely used datasets where classification tasks have $>5$K classes \cite{kuznetsova2020open, sun2017revisiting} while the popular object detection/segmentation datasets, like MSCOCO \cite{lin2014microsoft}, have annotations for only $80$ classes.
A similar observation was initially made by Hoffman \etal~\cite{hoffman2014lsda} and other semi-supervised \cite{kumar2018dock,tang2016large,uijlings2018revisiting} approaches.
%  that transform image-level classifiers into object detectors for {\em novel} classes by transferring knowledge from {\em base} classes. 
Second, the assumption of no instance-level supervision for target classes
% that no instance-level supervision is available for target classes 
(as is the case for semi-supervised \cite{hoffman2014lsda, kumar2018dock, tang2016large,uijlings2018revisiting} and zero-shot methods \cite{bansal2018zero, frome2013devise, rahman2018zero, xian2018zero}) is artificial. 
In practice, it is often easy to collect few instance-level annotations and, in general, a good object detection/segmentation model should be robust and work with {\em any} amount of available instance-level supervision. 
% \RG{NOTE-RCNN \cite{gao2019note} used this supervision in a training-mining setup \cite{tang2017multiple, uijlings2018revisiting} to refine detectors for target classes, and employed careful loss formulation and ensemble of detectors to regularize their approach at the cost of being simple and intuitive.} 
%Our motivation is to bridge weakly-supervised, zero- and few-shot learning paradigms to build an expressive, simple, and interpretable model that can operate and generalize with a type (weak/strong) and variety of instance-level supervision data (from $0$ to $30+$ \RG{(30?)} instance-level samples per class). 
%\RG{(`target' and `novel' are interchangeably used in the above para. Need to check whether this is ok.)}
Our motivation is to bridge weakly-supervised, zero- and few-shot learning paradigms to build an expressive, simple, and interpretable model that can operate across types (weak/strong) and amounts of instance-level supervision (from $0$ to $90+$ instance-level samples per class).

In this work, we develop a unified semi-supervised framework (UniT) for object detection and segmentation that scales with different levels of instance-level supervision ranging from no-data, to a few, to fully supervised (see Figure~\ref{fig:intro}). The data used in our problem is categorized in two ways, (1) image-level classification data for \emph{all} the object classes, and (2) abundant detection data for a set of \emph{base} object classes and limited (possibly zero) detection data for a set of \emph{novel} object classes, with the aim to obtain a model that learns to detect both \emph{base} and \emph{novel} objects at test time. %\sout{We note that this setting shares some similarity with few-shot object detection} \cite{Kang_2019_ICCV, Wang_2019_ICCV, Yan_2019_ICCV}, 

Our algorithm, illustrated in Figure~\ref{fig:model_arch}, jointly learns weak-detectors for {\em all} the object classes, from image-level classification data, and supervised regressors/segmentors on top of those for {\em base} classes (based on instance-level annotations in a supervised manner). The classifiers, regressors and segmentors of the {\em novel} classes are expressed as a weighted linear combination of its base class counterparts. The weights of the combination are determined by a multi-modal similarity measure: \emph{lingual} and \emph{visual}.
% The lingual similarity uses GloVe embeddings \cite{pennington2014glove} of class labels, while for visual similarity \sout{we leverage meta-learning} \RG{(need to modify this)}.
The key insight of our approach is to utilize the multi-modal similarity measure between the novel and base classes to enable effective knowledge transfer and adaptation. 
The adopted {\em novel} classifier/regressors/segmentors can further be refined based on instance-level supervision, if any available.  
We experiment with the widely-used detection/segmentation datasets - Pascal VOC \cite{everingham2015pascal} and MSCOCO \cite{lin2014microsoft}, and compare our method with state-of-the-art few-shot, weakly-supervised, and semi-supervised object detection/segmentation methods.

\vspace{0.07in}
\noindent
{\bf Contributions:}
%Our contributions can be summarized as follows: (1) We study the problem of weakly-supervised object detection (image-level annotation) in light of limited detection/segmentation data ranging from no data (zero-shot) to a few (few-shot) % , to fully 
%supervised data regimes; \RG{(unified)} (2) We propose a general, semantic and flexible end-to-end framework that can adopt classifiers/detectors/segmentors for {\em novel} classes by expressing them as linear combinations of their {\em base} class counterparts. In doing so, we leverage a learned multi-modal (lingual + visual) similarity metric. \RG{(stress: adopting regressors and segmentors)}
%\RG{(3) annotation budget claim} (4) We illustrate flexibility and effectiveness of our model by applying it to a variety of tasks (object detection and segmentation) and datasets (Pascal VOC \cite{everingham2015pascal}, MSCOCO \cite{lin2014microsoft}); showing state-of-the-art performance. On MS-COCO, we get as much as $+7.5$/$+10.4$ mAP on detection/segmentation over the closest baseline 
Our contributions can be summarized as follows: (1) We study the problem of semi-supervised object detection and segmentation in light of image-level supervision and limited instance-level annotations, ranging from no data (zero-shot) to a few (few-shot); (2) We propose a general, unified, interpretable and flexible end-to-end framework that can adopt classifiers/detectors/segmentors for {\em novel} classes by expressing them as linear combinations of their {\em base} class counterparts. In doing so, we leverage a learned multi-modal (lingual + visual) similarity metric. (3) In the context of our model, we explore the relative importance of weak image-level supervision, compared to strong instance-level supervision, and find that under a small fixed annotation budget, image-level supervision is more important. (4) We illustrate flexibility and effectiveness of our model by applying it to a variety of tasks (object detection and segmentation) and datasets (Pascal VOC \cite{everingham2015pascal}, MSCOCO \cite{lin2014microsoft}); showing state-of-the-art performance. We get up to $23\%$ relative improvement in mAP over the closest semi-supervised methods \cite{gao2019note}, and as much as $16\%$ improvement with respect the best performing few-shot method \cite{wang2020frustratingly} under the same fixed annotation budget.
% On MS-COCO, we get as much as $+7.5$/$+10.4$ mAP on detection/segmentation over the closest baseline \cite{Yan_2019_ICCV}. 
%To the best of our knowledge ours is the first paper that conducts comprehensive comparisons across settings, tasks, types and levels of supervision.  
We conduct comprehensive comparisons across settings, tasks, types and levels of supervision.

\vspace{-0.07in}
%our improvement is as much as 7.5/10.4 mAP on detection/segmentation over the closest baseline \cite{Yan_2019_ICCV}. % and 10.4 mAP on segmentation over the closest baseline \cite{Yan_2019_ICCV}. 

% Our contributions can be summarized as follows:
% \vspace{-0.08in}
% \begin{enumerate}
%     \item We study the problem of weakly supervised object detection (image-level annotation) in the light of limited detection data ranging from no data (zero-shot), to a few (few-shot), to fully supervised (transfer) data regimes.
%     \item We design a general framework that learns and adapt in any of the above limited data settings ...
%     \item We observe and experimentally show that classifiers and detectors of novel classes can be semantically decomposed into their base classes' counterparts.
% \end{enumerate}

\section{Related Work}
\vspace{-0.05in}
\noindent
{\bf Few-shot object detection:}
Object detection with limited data was initially explored in a transfer learning setting by Chen \etal~\cite{chen2018lstd}. 
% which has the potential to overfit. 
% \sout{Lately, meta-learning} \cite{andrychowicz2016learning, ravi2016optimization, finn2017model, snell2017prototypical, vinyals2016matching} \sout{has emerged as a paradigm which attempts to resolve overfitting by simulating a \emph{learning-to-learn} scheme with episodic tasks}. 
In the context of % \sout{object detection} 
meta-learning \cite{andrychowicz2016learning, finn2017model, ravi2016optimization, snell2017prototypical, vinyals2016matching}, Kang \etal~\cite{Kang_2019_ICCV} 
% put forward 
developed a few-shot model where the learning procedure is divided into two phases: first the model is trained on a set of \emph{base} classes with abundant data using episodic tasks, then, in the second phase, a few examples of \emph{novel} classes and \emph{base} classes are used for fine tuning the model. Following this formulation, \cite{Wang_2019_ICCV, Yan_2019_ICCV} employed better performing architecture - Faster R-CNN \cite{ren2015faster}, instead of a one-stage YOLOv2 \cite{redmon2017yolo9000}. Yan \etal~\cite{Yan_2019_ICCV} 
% also
extended the problem formulation to account for segmentation 
% masks 
in addition to detection. 
% However, contrary 
In contrast to the above approaches, Wang \etal~\cite{wang2020frustratingly} showed that meta-learning is not a crucial ingredient to Few-shot object detection, and simple fine-tuning produces better detectors.
Similar to the above works, we also adopt the two-phase learning procedure. % used in few-shot object detection \cite{Kang_2019_ICCV,Yan_2019_ICCV, Wang_2019_ICCV, wang2020frustratingly}. 
However, we fundamentally differ in assuming that easily attainable extra supervision, in the form of image-level data, over \emph{all} the classes is available. Unlike \cite{Wang_2019_ICCV}, we learn a semantic mapping between {\em weakly-supervised detectors} and detectors obtained using a large number of examples.

\vspace{0.03in}
\noindent
{\bf Weakly-supervised object detection:} Weak supervision in object detection takes the form of image-level labels, usually coupled with % externally pre-computed 
bounding box proposals \cite{uijlings2013selective, zitnick2014edge}, thereby representing each image as a bag of instances \cite{arun2019dissimilarity, bilen2016weakly, cinbis2016weakly, diba2017weakly, gao2019c, ren2020instance, song2014weakly, tang2018pcl, tang2017multiple, wang2018collaborative, zeng2019wsod2}. Bilen \etal~\cite{bilen2016weakly} proposed an end-to-end architecture which softly labeled object proposals and uses a detection stream, in addition to classification stream, to classify them. Further extensions followed, Diba \etal~\cite{diba2017weakly} incorporated better proposals into a cascaded deep network; Tang \etal~\cite{tang2017multiple} proposed an Online Instance Classifier Refinement (OICR) algorithm which iteratively refines predictions. % by propagating the instance predictions to their spatially-overlapping instances. 
More recently, further improvements were made by combining weakly-supervised learning with strongly-supervised detectors, by treating predicted locations from the weakly-supervised detector as pseudo-labels for a % the learning of the 
strongly-supervised variant \cite{arun2019dissimilarity, wang2018collaborative}. 
% \RG{TO DO: Check SOTA - Wetectron \cite{ren2020instance} and other close performers - WSOD2 \cite{zeng2019wsod2} and C-MIDN \cite{gao2019c}, and see if anything can be said here.}
% , and our currently state-of-the-art approaches. 
% We compare our approach with these methods in section ().
In this work, we choose to adopt and build on top of single-stage OICR \cite{tang2017multiple}, %  which is relatively simple and single-stage, 
hence enabling end-to-end training. However, our approach is not limited to the choice of weakly-supervised architecture.
\vspace{0.03in}
\noindent
{\bf Semi-supervised object detection:}
Approaches under semi-supervised setup assume abundant detection data for {\em base} classes and no detection data for {\em novel} classes, in addition to weak supervision for {\em all} the classes. The methods in this category first learn weak classifiers for {\em all} classes using abundant weak supervision, then fine-tune {\em base} classifiers into detectors using abundant detection data, and finally transfer this transformation to obtain detectors for {\em novel} classes using an external (or learned) similarity measure between {\em base} and {\em novel} classes. % The methods in this category aim to transfer knowledge from {\em base} classes to {\em novel} classes using an external or learned similarity measure, and adapt weak classifiers of novel classes to obtain their final detectors.
LSDA \cite{hoffman2014lsda}, being the first,  formed similarity based on L2-normalized weak classifier weights.
%Notably, LSDA \cite{hoffman2014lsda} is the first attempt in this direction where they formed similarity based on L2-normalized weak classifier weights. 
Tang \etal~\cite{tang2016large} extended this approach to include semantic and visual similarity % components 
explicitly. DOCK \cite{kumar2018dock} expanded the types of similarities to include spatial and attribute cues using external knowledge sources. Other works leverage semantic hierarchies of classes, such as Yang \etal~\cite{yang2019detecting} proposes a class split based on granularity of classes, and transfers knowledge from coarse to fine grained classes. Uijlings \etal~\cite{uijlings2018revisiting} uses a proposal generator trained on base classes, and transfers the
proposals from base to novel classes by computing their  similarity on a tree based on Imagenet semantic hierarchy \cite{russakovsky2015imagenet}. % proposal scoring function based on the desired level of generality for novel classes, ranging from class-generic `entities' to class-specific categories such as `tiger' and `cat' (rephrase).
Similar to the above methods we also use % the canonical 
visual and lingual similarities between base and novel classes,
% However, we 
but consider a more general problem setting where we have varying degrees of detection supervision for novel classes ranging from zero to a few $k$-samples per class. 
% As such, the setup used in the above methods can be seen as a simplified variant of our model with $k=0$. \RG{TO DO: Check \cite{hoffman2015detector}} \Leon{Can we say more about our distinction?}

Unique, and closest to our setup, is NOTE-RCNN \cite{gao2019note}. In \cite{gao2019note}, 
% where they termed 
few-$k$ detection samples for {\em novel} classes are used as seed annotations, based on which training-mining \cite{tang2017multiple, uijlings2018revisiting} is employed. 
%  . They employ a training-mining setup \cite{tang2017multiple, uijlings2018revisiting}, where 
Specifically, they initialize detectors for {\em novel} classes by training them with few seed annotations,
% then the detectors are refined by 
and iteratively refine them by retraining with mined bounding boxes for novel classes. They transfer knowledge indirectly in the form of losses that act as regularizers. %, and use an ensemble of detectors to obtain predictions. 
Our approach, on the other hand, takes on a simpler and more intuitive direction where we first transfer the mappings from {\em base} to {\em novel} classes, and use few seed annotations (if available) to fine-tune the detectors. 
Despite being simpler, our approach is more accurate, and works in the $k=0$ regime. 
Further, unlike all the above semi-supervised approaches
% , including \cite{gao2019note}
, we transfer across tasks, including regression and segmentation.  %, and not just classification.

% This simpler approach has been shown to be effective \cite{wang2020frustratingly}, and has a desirable property of being deployed as a single framework for varying degrees of supervision including none.

% Notably, LSDA \cite{hoffman2014lsda} assumes image-level supervision for all the classes and detection supervision for a subset of those classes ({\em base}). Their approach adapts image-level classifiers to object detectors by learning a transformation on {\em base} classes, with the aim to transfer this knowledge to {\em novel} classes with no detection data. 

% In contrast, our model starts with weakly-supervised detectors, instead of classifiers,  and considers a more generalized problem setting where we have varying degrees of detection supervision for novel classes ranging from zero to a few $k$-samples per class. As such, the setting in LSDA is simplified variant of our model with $k=0$. 

\vspace{0.03in}
\noindent
{\bf Zero-shot object detection:}
Zero-shot approaches 
% This setup assumes no data for the novel classes, so methods in this category 
rely on auxiliary semantic information to connect {\em base} and {\em novel} classes; {\em e.g.}, text description of object labels or their attributes \cite{bansal2018zero, frome2013devise, rahman2018zero, xian2018zero}. 
A common strategy is to represent {\em all} classes as prototypes in the semantic embedding space and to learn a mapping from visual features to this embedding space using {\em base} class data; classification is then obtained using nearest distance to {\em novel} prototypes. This approach was expended to detection in \cite{demirel2018zero, kuznetsova2015expanding, li2019zero, sadhu2019zero, zablocki2019context, zhu2019dont}.
% A common theme among these methods is to learn a compatibility function between visual features and embeddings of object labels (or attributes), and during evaluation predict the class which attains the maximum compatibility score with the visual features of a test image. 
Bansal \etal~\cite{bansal2018zero}, similarly, 
% extended the above framework for object detection and in particular 
proposed method to deal with situations where objects from novel/unseen classes are present in the background regions. We too explore the setting where we are not provided with any instance data for novel classes, but in addition assume weak-supervision for novel object classes in the form of readily available \cite{kuznetsova2020open} image-level annotations.

\vspace{-0.06in}
\section{Problem Formulation}
\vspace{-0.06in}

Here we formally introduce the semi-supervised any-shot object detection / segmentation setup. We start by assuming image-level supervision for \emph{all} the classes denoted by $\classdata = \{(\mathbf{x}_i, \mathbf{a}_i)\}$, where each image $\mathbf{x}_i$ is annotated with a label $\mathbf{a}_i \in \{0, 1\}^{|\mathcal{C}|}$, where $a_i^{j} = 1$ if image $\mathbf{x}_i$ contains at least one $j$-th object, indicating its presence; $\mathbf{a}_i = \{ a_i^{j} \}_{j=1}^{|\mathcal{C}|}$ with $|\mathcal{C}|$ being number of object classes.

We further extend the above image-level data with object-instance annotations by following the few-shot object detection formulation \cite{Kang_2019_ICCV, Wang_2019_ICCV, Yan_2019_ICCV}. We split the classes into two disjoint sets: {\em base} classes $\bdata$ and {\em novel} classes $\ndata$; $\bdata \cap \ndata = \emptyset$. For base classes, we have abundant instance data $\bdetdata = \{(\mathbf{x}_i, \mathbf{c}_i, \mathbf{y}_i)\}$, where $\mathbf{x}_i$ is an input image, $\mathbf{c}_i = \{ c_{i,j} \}$ are class labels,  $\mathbf{y}_i = \{ \mathbf{bbox}_{i,j} \}$ or $\mathbf{y}_i = \{ \mathbf{s}_{i,j} \}$ are corresponding bounding boxes and/or masks for each instance $j$ in image $i$. 
% Whereas 
For {\em novel} classes, we have limited instance data $\ndetdata = \{(\mathbf{x}_i, \mathbf{c}_i, \mathbf{y}_i)\}$, where a $k$-shot detection / segmentation data only has $k$ bounding boxes / masks for each novel class in $\ndata$. 
Note, for semi-supervised zero-shot, $k=0$ and $\ndetdata = \emptyset$. 

\vspace{-0.08in}
\section{Approach}
\label{sec:approach}
\vspace{-0.06in}
\begin{figure*}[]
    \centering
    \includegraphics[width=0.8\textwidth]{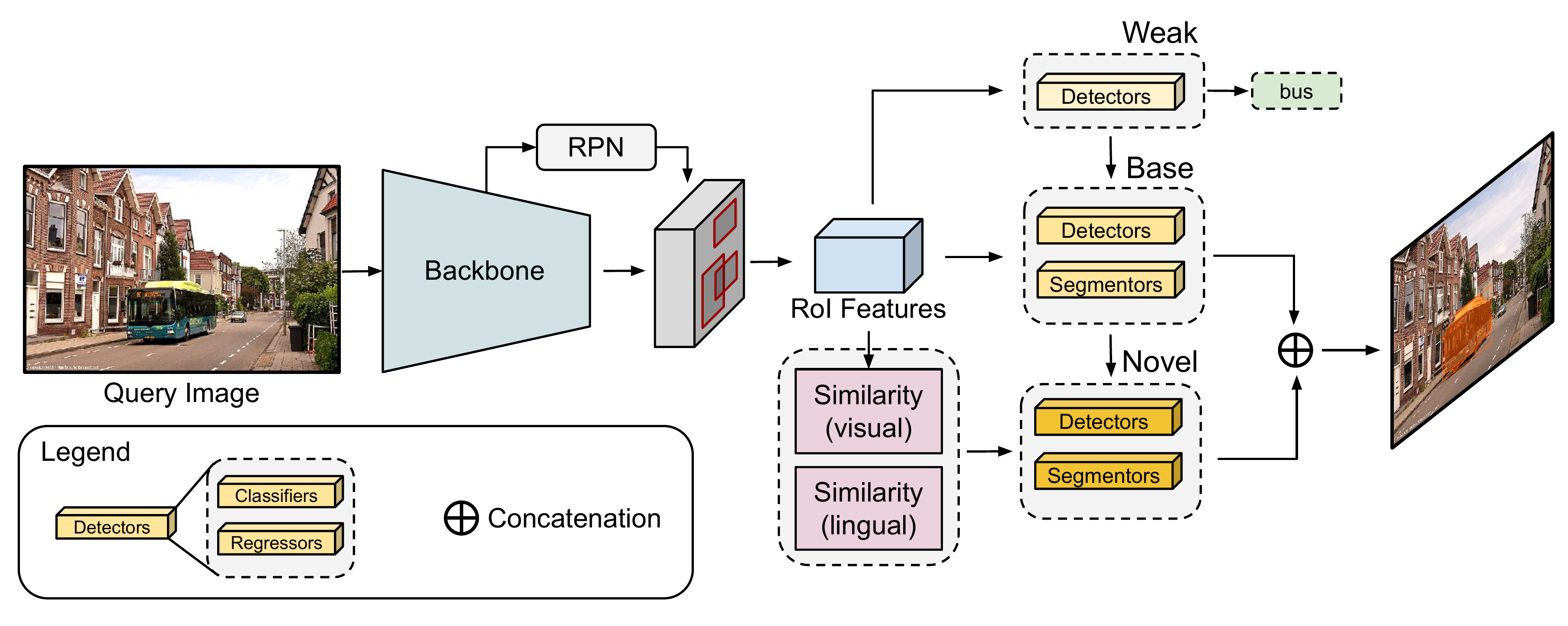}
    \vspace{-0.2in}
    \caption{{\bf Overall Architecture.} 
    We form detectors/segmentors of {\em base} classes as a refinement on top of the weak detectors. The detectors/segmentors of {\em novel} classes utilize a similarity weighed transfer from the base class  refinements. 
    In $k$-shot setting, (few) novel class instance  annotations % additional samples 
    are incorporated through direct adaptation of the  resulting {\em novel} detectors/segmenters through fine-tuning.
    The similarity is a combination of {\em lingual} and {\em visual} similarity (pink boxes). 
    All detectors are built on top of Faster/Mask RCNN architecture which comprises of classification and regression heads with shared backbone (in cyan) and simultaneously trained region proposal network (RPN).} 
    % All components are trained end-to-end.}
    % We form detectors/segmentors of novel classes as a function of weak detectors, detectors/segmentors of base classes, and similarity (lingual and visual) matrices.
    % \RG{For some reason, the legend thing looks off. Let me know your comments.}}
    \label{fig:model_arch}
    \vspace{-0.26in}
\end{figure*}
We propose a single unified framework that leverages the weak image-level supervision for object detection / segmentation in any-shot setting. That is, our proposed approach can seamlessly incorporate arbitrary levels of instance-level supervision without the need to % significantly 
alter the architecture. 

% Due to their impressive performance on detection / segmentation tasks, 
Our proposed framework builds upon the Faster R-CNN \cite{ren2015faster} / Mask R-CNN \cite{he2017mask} architecture. Faster R-CNN \cite{ren2015faster} utilizes a two-stage pipeline in order to perform object detection. The first stage uses a region proposal network (RPN) to generate class-agnostic object region proposals $\{ \mathbf{rbox}_{i,j} \}$ for image $i$. The second stage is a detection network (Fast R-CNN \cite{girshick2015fast}) that performs RoI pooling, forming feature vector $\mathbf{z}_{i,j} = \text{\tt RoIAlign}(\mathbf{x}_i, \mathbf{rbox}_{i,j})$ for proposal $j$ in image $i$, and learns to classify this RoI feature vector $\mathbf{z}$ (we drop proposal and image indexing for brevity for remainder of the section) into one of the object classes and refine the bounding box proposals using a class-aware regressors. Conceptually, an R-CNN object detector can be thought of as a combination of a classifier and regressor (see Figure \ref{fig:model_arch}). Mask R-CNN \cite{he2017mask} is a simple extension to the Faster R-CNN framework, wherein an additional head is utilized in the second stage to predict the instance segmentation masks.

% \RG{For the following para: \\
% - Two branches of our approach and second stage of Faster-RCNN in one paragraph (could create confusion) \\
% - Mention that weak detector consists of a classifier and 0 regressor. Potentially change the equations to reflect that or whatever we discussed. \\
% - Mention that (by the virtue of) joint training, it enables us to train the weak branch by using proposals from the strong/supervised branch.
% }

Figure \ref{fig:model_arch} details the proposed architecture. The model consists of two branches: i) the weakly-supervised branch that trains detectors $\hat{c} = \text{\tt softmax}(f_{\mathbf{\W}^{weak}}(\mathbf{z}))$ using image-level supervision $\classdata$, and ii) a supervised branch that uses detection data $\bdetdata$/$\ndetdata$ to learn a refinement mapping from the weak detector to category-aware classifiers, regressors, and segmentors $f_{\mathbf{\W}^{*}}(\mathbf{z}); * \in \{cls, reg, seg\}$, which are used in the second stage of Faster / Mask R-CNN. 
Note that weak detectors simply output the proposal box of the pooled feature vector as the final location $\hat{\mathbf{y}} = \mathbf{rbox}$; 
%\Sid{(changed pbox to rbox as that's what we define in approach)}, 
while refined detectors are able to regress a better box. %  $\hat{\mathbf{y}}  = \mathbf{pbox} + f_{\mathbf{\W}^{reg}}(\mathbf{z})$.
Here $f_{\mathbf{\W}}(\cdot)$ is a learned neural network function parametrized by $\mathbf{\W}$. 
% Additionally, all the functions $f$ mentioned in the paper operate over proposals $\mathbf{z}_{i,j}$, which is omitted for brevity.
% These detectors serve as the initialization (or starting point) for the final detection / segmentation weights.
% The second branch uses detection data $\bdetdata$/$\ndetdata$ to learn a mapping over $\mathbf{\W}^{weak}$ to transform these image-level detectors into final detectors.
We jointly train both branches and the RPN, and 
% since our approach follows the meta-learning paradigm, 
learning is divided into two stages: base-training and fine-tuning\footnote{We use the nomenclature introduced in \cite{Kang_2019_ICCV}.}. % meta-training and meta-testing. 

\vspace{0.1em}
\noindent{\bf Base-training:} During base-training, instances from $\bdetdata$ are used to obtain a detector / segmentation network for the {\em base} classes $\bdata$. 
Specifically, for each $b \in \bdata$, category-aware classifiers and regressors for the base classes are formulated as additive refinements to their corresponding weak counterparts. For region classifiers this takes the form of: $\hat{c} = \underset{\bdata}{\argmax}\, \left[ \text{\small\tt softmax} \left(f_{\mathbf{\W}_{base}^{cls}}(\mathbf{z}) \right) \right]$, where %  as  follows: % $\mathbf{\W}_{base}^{cls}$ as follows:
\begin{equation}
    %\hat{c} = \underset{\bdata}{\argmax}\, \left[ \text{\tt softmax} \left(f_{\mathbf{\W}_{base}^{cls}}(\mathbf{z}) \right) \right], ~~~~~~~~
    f_{\mathbf{\W}_{base}^{cls}}(\mathbf{z}) = f_{\mathbf{\W}_{base}^{weak}}(\mathbf{z}) + f_{\Delta\mathbf{\W}_{base}^{cls}}(\mathbf{z}), 
    \label{eq:cls-base}
\end{equation}
where $f_{\Delta\mathbf{\W}_{base}^{cls}}(\mathbf{z})$ is a zero-initialized residual to the logits of the weakly supervised detector.  % $f(\mathbf{\W}_{weak}^{cls})$. 
The regressed object location is similarly defined as: 
\begin{equation}
      \hat{\mathbf{y}} =\mathbf{rbox} + f_{\mathbf{\W}_{base}^{reg}}(\mathbf{z}).
      \label{eq:reg-base}
\end{equation}
% , and $\sigma$ is the softmax non-linearity. 
% Note that ${\W}_{b,base}^{reg}$ and ${\W}_{b,base}^{seg}$ are estimated using the standard Faster / Mask R-CNN learning procedure.
% The segmentor can be defined analogously and is omitted due to lack of space (see Suppl. Materials). 
Finally, as there is no estimate for the segmentation masks in the first stage of Mask R-CNN \cite{he2017mask}, $\hat{\mathbf{y}} = f_{\mathbf{\W}_{base}^{seg}} (\mathbf{z})$ is learned directly from the \emph{base} annotations.
% The segmentor is define analogously to the regressor. Please see Section A of the supplementary for additional details.

\vspace{0.1em}
\noindent{\bf Novel fine-tuning  ($k > 0$):} 
In the fine-tuning phase, the detectors / segmentors of the base classes are used to transfer information to the classes in $\ndata$. 
The network is also fine-tuned on $\ndetdata$, which, for a value of $k$, contains $k$ bounding boxes / masks for novel and base classes. 
Here we consider the case of $k > 0$;  we later address $k = 0$ case, which does not require fine-tuning. 
The key insight of our approach is to use additional \emph{visual} and \emph{lingual} similarities between the {\em novel} and {\em base} classes 
to enable effective transfer of the network onto the {\em novel} classes under varying degrees of supervision. Contrary to existing work \cite{hoffman2014lsda,tang2016large,kumar2018dock} that only consider information from {\em base} category-aware classifiers, our approach additionally learns a mapping from {\em base} category-aware regressors and segmentors to obtain more accurate \emph{novel} counterparts.
For a specific proposal $\mathbf{rbox}$ with features $\mathbf{z}$, let $\mathbf{S}(\mathbf{z}) \in  \mathbb{R}^{|\ndata| \times |\bdata|}$ 
denote similarity between base classes and novel classes.  
The dependence on $\mathbf{z}$ stems from visual component of the similarity and is discussed in Section \ref{sec:similarity}.
% 
% $\mathbf{S}^{lin}_{n,b} \in  \mathbb{R}^{|\ndata| \times |\bdata|}$ and $\mathbf{S}^{vis}_{b}(\mathbf{z}) \in  \mathbb{R}^{|\bdata|}$ denote the lingual and visual similarities respectively. 
%  The details for computing these similarities are discussed in Section \ref{sec:similarity}. 
Given this, for each proposal $\mathbf{z}$, the category-aware classifier for the novel classes is obtained as follows: $\hat{c} = \underset{\ndata}{\argmax}\, \left[ \text{\tt softmax} \left(f_{\mathbf{\W}_{novel}^{cls}}(\mathbf{z}) \right) \right]$, where $f_{\mathbf{\W}_{novel}^{cls}}(\mathbf{z})$ can be written as,
\begin{align}
\begin{split}
% \scriptsize
     %\hat{c} = \underset{\ndata}{\argmax}\, \left[ \text{\tt softmax} \left(f_{\mathbf{\W}_{novel}^{cls}}(\mathbf{z}) \right) \right], ~~~~~~~~
    % &f_{\mathbf{\W}_{novel}^{cls}}(\mathbf{z}) = \\
    &\underbrace{f_{\mathbf{\W}_{novel}^{weak}}(\mathbf{z})}_{\substack{\text{weak-detectors}}} + \underbrace{\mathbf{S}(\mathbf{z})^T f_{\Delta \mathbf{\W}_{base}^{cls}}(\mathbf{z})}_{\substack{\text{instance-level transfer} \\ \text{from base classes}}} +
    \underbrace{f_{\Delta \mathbf{W}_{novel}^{cls}}(\mathbf{z})}_{\substack{\text{instance-level} \\ \text{direct adaptation}}}
    \label{eq:cls}
\end{split}
\end{align}
where $\mathbf{S}(\mathbf{z}) = \text{\tt softmax}(\mathbf{S}^{lin} \odot \mathbf{S}^{vis}(\mathbf{z}))$, and $\odot$ denotes broadcast of vector similarity $\mathbf{S}^{vis}(\mathbf{z}) \in \mathbb{R}^{|\bdata|}$ followed by element-wise product with lingual similarity  $\mathbf{S}^{lin} \in  \mathbb{R}^{|\ndata| \times |\bdata|}$. 
The interpretation of Eq.(\ref{eq:cls}) is actually rather simple  -- we first refine the weak detectors for novel classes by similarity weighted additive refinements from base classes  ({\em e.g.}, novel class {\tt motorbike} may relay on base class {\tt bicycle} for refinement; illustrations in Section \ref{sec:similaritymatrix} of the appendix.), denoted by ``instance-level transfer from base classes'';
% \footnote{Note that the only learnable parameter is the visual component of similarity matrix $\mathbf{S}(\mathbf{z})$.};
we then further directly adapt the resulting detector (last term) with few instances of the novel class. 
Similarly, for each $\mathbf{z}$, the novel class object regressor can be obtained as,
\begin{align}
\begin{split}
    \hat{\mathbf{y}} &= \mathbf{rbox} + 
    f_{\mathbf{\W}_{novel}^{reg}}(\mathbf{z}) \\
    &= \mathbf{rbox} + 
    \underbrace{\mathbf{S}^T(\mathbf{z}) f_{\mathbf{\W}_{base}^{reg}}(\mathbf{z})}_{\substack{\text{instance-level transfer} \\ \text{from base classes}}} +
    \underbrace{f_{\Delta \mathbf{\W}_{novel}^{reg}}(\mathbf{z})}_{\substack{\text{instance-level} \\ \text{direct adaptation}}}
    \label{eq:reg}
\end{split}
\end{align}
Finally, the segmentation head $f_{\mathbf{\W}_{novel}^{seg}}(\mathbf{z})$ can be obtained as follows (additional details in appendix Section \ref{sec:segmentation}),
\begin{align}
    \hat{\mathbf{y}} = f_{\mathbf{\W}_{novel}^{seg}}(\mathbf{z}) =  
    \underbrace{\mathbf{S}^T(\mathbf{z}) f_{\mathbf{\W}_{base}^{seg}}(\mathbf{z})}_{\substack{\text{instance-level transfer} \\ \text{from base classes}}} +
    \underbrace{f_{\Delta \mathbf{\W}_{novel}^{seg}}(\mathbf{z})}_{\substack{\text{instance-level} \\ \text{direct adaptation}}}
    \label{eq:seg}
\end{align}
% identical to Eq.(\ref{eq:reg}). In the following sections we describe the individual elements in more detail.

% \vspace{0.4em}
\noindent{\bf Semi-supervised zero-shot ($k\!=\!0$):} 
As we mentioned previously, our model is also readily applicable when $\ndata = \emptyset$. 
This is a special case of the formulation above, where fine-tuning is not necessary or possible, and we only rely on base training and apply novel class evaluation procedure. 
The predictions for novel classes can be done as in Eq.(\ref{eq:cls}), Eq.(\ref{eq:reg}), and Eq.(\ref{eq:seg}), but omitting the ``instance-level direct adaptation'' term in all three cases.

%  \RG{Mention $k=0$ case (the direct adaptation term disappears) and how our approach is flexible to deal with it.}

\vspace{-0.06in}
\subsection{Weakly-Supervised Detector}
\vspace{-0.06in}
As mentioned earlier, our approach leverages detectors trained on image level annotations to learn a mapping to supervised detectors/segmentors. We highlight that our approach is agnostic to the method used to train the weakly-supervised detector, and most of the existing approaches \cite{arun2019dissimilarity, bilen2016weakly, tang2018pcl,  tang2017multiple} can be integrated into our framework. We, however, use the Online Instance Classifer Refinemnet (OICR) architecture proposed by Tang \etal~\cite{tang2017multiple} due to its simple architecture. OICR has $R$ ``refinement'' modules $f_{\mathbf{W}^{weak}_r}(\mathbf{z})$ that progressively improve the detection quality. These individual ``refinement'' modules are combined to obtain the final prediction as follows,
\begin{equation}
    \hat{\mathbf{a}} = \text{\small\tt softmax} \left[ f_{\mathbf{W}^{weak}}(\mathbf{z}) \right] = \text{\small\tt softmax} \left[ \frac{1}{R}\sum_{r} f_{\mathbf{W}^{weak}_r}(\mathbf{z}) \right]
    \label{eq:weak-labels}
\end{equation}
We use the same loss formulation $\mathcal{L}^{weak}(\mathbf{a}, \hat{\mathbf{a}})$ described in \cite{tang2017multiple}, which compares predicted ($\hat{\mathbf{a}}$) and ground truth ($\mathbf{a}$) class labels,
to train the OICR module (see Sect. \ref{sec:loss}). For additional details, we refer the reader to %  , we would point the reader to 
\cite{tang2017multiple}. 

% We adopt the ResNet \cite{he2016deep} architecture, pre-trained on the ImageNet-1k \cite{imagenet_cvpr09} dataset, to obtain our classifier $f(\classdata;\boldsymbol{\theta})$ on weak image-level data $\classdata$. Using an ImageNet-1k pre-trained network not only helps with performance, but also allows easy adaptation to a new domain with limited data \cite{girshick2014rich, russakovsky2015imagenet}. We replace the last weight layer (1000 linear classifiers) with $|\bdata \cup \ndata|$ linear classifiers, and fine-tune the entire network on \emph{image-level} annotations (class labels) in $\classdata$. Note that we don't use any bounding box / segmentation masks annotations during this training procedure. $f(\classdata;\boldsymbol{\theta})$ is used to initialize our detector / segmentation model.

\subsection{Similarity Matrices}\label{sec:similarity}
\vspace{-0.06in}
As described in Eq.(\ref{eq:cls}), (\ref{eq:reg}), (\ref{eq:seg}), the key contribution of our approach is the ability to semantically decompose the classifiers, detectors and segmentors of novel classes into their base classes' counterparts. To this end, we define a proposal-aware similarity $\mathbf{S}(\mathbf{z}) \in \mathbb{R}^{|\ndata| \times |\bdata|}$, where each element captures the semantic similarity of novel class $n$ to base class $b$. We assume the similarity matrix $\mathbf{S}(\mathbf{z})$ can be decomposed into two components: \emph{lingual} % similarity 
$\mathbf{S}^{lin}$ and \emph{visual} $\mathbf{S}^{vis}(\mathbf{z})$ similarity. % $\mathbf{S}^{vis}(\mathbf{z})$. Let $\mathbf{S}^{*}_{n,b}; * \in \{lin, vis\}$ denote the $(n,b)$-th entry of the corresponding similarity array. 

\vspace{0.04in}
\noindent
\textbf{Lingual Similarity:} This term captures linguistic similarity between novel and base class labels. The intuition lies in the observation that semantically similar classes often have correlated occurrences in textual data. Therefore, for a novel class $n$ and a base class $b$, $\mathbf{S}_{n,b}^{lin} = \mathbf{g}_n^\top \mathbf{g}_b$, where $\mathbf{g}_n$ and $\mathbf{g}_b$ are 300-dimensional GloVe \cite{pennington2014glove} vector embeddings for $n$ and $b$ respectively\footnote{For class names that contain multiple words, we average individual GloVe word embeddings.}.

% A $k$-shot task consists of a \emph{support set} $\supportset = \{(\mathbf{x}_i, \mathbf{c}_i, \mathbf{y}_i)\}$ and a \emph{query set} $\queryset = \{(\mathbf{x}_i^*, \mathbf{c}_j^*, \mathbf{y}_i^*)\}$. The support set $\supportset$ contains $k$ images for each class, where each image is annotated with an object bounding box / mask belonging to the corresponding class. The query set $\queryset$ contains images with annotations for evaluation.

\vspace{0.04in}
\noindent
\textbf{Visual Similarity: } Complementary to the lingual component, this \emph{proposal-aware} similarity models the visual likeness of a proposal $\mathbf{z}$ to objects from the \emph{base} classes. To this end, for each $\mathbf{z}$, we use the normalized predictions $\hat{\mathbf{a}}$ of the weak detector $f_{\mathbf{W}^{weak}}(\mathbf{z})$, as described in Eq. (\ref{eq:weak-labels}), as a proxy for the likelihood of $\mathbf{z}$ belonging to a \emph{base} class $b$. Specifically, let $\hat{\mathbf{a}}_b$ be the score corresponding to the base class $b$. For a novel class $n$ and a base class $b$, the visual similarity $\mathbf{S}^{vis}_{n,b}(\mathbf{z})$ is then defined as,
\begin{equation}
    \mathbf{S}^{vis}_{n,b}(\mathbf{z}) = \frac{\hat{\mathbf{a}}_b}{\sum_b \hat{\mathbf{a}}_b}
\end{equation}
Note that computing this visual similarity \emph{does not} require learning additional parameters. Rather, it is just a convenient by-product of training our model. As a result, this similarity can be efficiently computed. Our proposed formulation for visual similarity, in its essence, is similar to the one used in \cite{tang2016large}.
% , wang2020frustratingly}. 
However, \cite{tang2016large} use \emph{image-level} scores aggregated over validation set, lacking ability to adapt to a specific proposal. 
% Also, we find that computing a cosine similarity between $\mathbf{z}$ and the parameters of the category-aware classifier $f_{\mathbf{\W}_{base}^{cls}}$, as in \cite{wang2020frustratingly}, tends to \emph{over-estimate} the likelihood of $\mathbf{z}$ belonging to a base class $b$. This is in part due to the fact that $f_{\mathbf{\W}_{base}^{cls}}$ isn't trained to classify \emph{novel} classes correctly. Using the weak detector $f_{\mathbf{W}^{weak}}$ circumvents this issue.

We would also like to highlight that our framework is extremely flexible and can easily utilize any additional information, similar to \cite{kumar2018dock}, to obtain a more accurate semantic decomposition $\mathbf{S(z)}$. However, as computing these might require the use of additional datasets and pre-trained models, we refrain from incorporating them into our model. 

\vspace{-0.06in}
\subsection{Training}
\vspace{-0.06in}
\label{sec:loss}
We now describe the optimization objective used to train our proposed approach in an end-to-end fashion. 
% As mentioned earlier, due to the meta-learning nature of the task, we use separate objectives during base training and fine-tuning. 
During base training, the objective can be written as,
\begin{equation}
    \mathcal{L}^{t} = \mathcal{L}^{rcnn} + \alpha\mathcal{L}^{weak}
\end{equation}
where $\mathcal{L}^{rcnn}$ is the Faster/Mask R-CNN \cite{he2017mask, ren2015faster} objective, and $\mathcal{L}^{weak}$ is the OICR \cite{tang2017multiple} objective; $\alpha=1$ is the weighting hyperparameter. In fine-tuning, we refine the model only using  $\mathcal{L}^{rcnn}$. 
% We note that 
Fine-tuning only effects last term of Eq.(\ref{eq:cls}), (\ref{eq:reg}), and (\ref{eq:seg}), while everything else is optimized using base training objective. Further implementation details are in Section \ref{sec:implementation} of the appendix. We will  make all code and pre-trained models {\bf publicly available}.

\vspace{-0.06in}
\section{Experiments}
\vspace{-0.08in}
We evaluate our approach against related methods in the semi-supervised and few-shot domain. Comparison against work in the weakly-supervised object detection literature is provided in appendix Section \ref{exp:weaklyexp}.
\vspace{-0.04in}
% In this section, we provide experiments for evaluating our approach to related methods: few-shot object detection and segmentation, weakly-supervised object detection, including an ablation study.
\subsection{Semi-supervised Object Detection}
\label{exp:semisup}
\vspace{-0.08in}
\noindent\textbf{Datasets.} We evaluate the performance of our framework on MSCOCO \cite{lin2014microsoft} 2015 and 2017 datasets. Similar to \cite{gao2019note, kumar2018dock}, we divide the $80$ object categories into $20$ \emph{base} and $60$ \emph{novel} classes, where the \emph{base} classes are identical to the $20$ VOC \cite{everingham2010pascal} categories. For our model and the baselines, we assume image-level supervision for \emph{all} $80$
classes, whereas instance-level supervision is only available for $20$ \emph{base} classes. For few-shot experiments $(k > 0)$ we additionally assume $k$ instance-level annotations for the \emph{novel} classes. 
\vspace{0.3em}
% We evaluate on MSCOCO 2015 and 2017 \cite{lin2014microsoft} datasets similar to the previous works \cite{gao2019note, kumar2018dock}, and use the prescribed class split for $80$ object categories where $20$ VOC \cite{everingham2010pascal} categories are assigned as {\em base} classes and the remaining $60$ as the {\em novel} classes.
% Also we evaluate using the same evaluation metric where NOTE-RCNN uses the standard mAP \cite{lin2014microsoft} averaged over IoU thresholds in $[0.5: 0.05: 0.95]$ denoted as AP, and DOCK uses mAP at IoU threshold $0.5$ denoted as AP50.

\noindent\textbf{Semi-supervised zero-shot ($k\!=\!0$).} Table \ref{tab:semi-zero-shot} compares the performance of our proposed approach against the most relevant semi-supervised zero-shot ($k=0$) methods \cite{hoffman2014lsda, hoffman2015detector,kumar2018dock,tang2016large} on \emph{novel} classes. As an upper-bound, we also show the performance of a fully-supervised model. To ensure fair comparison, we follow the experimental setting in the strongest baseline DOCK \cite{kumar2018dock}, and borrow the performance for \cite{hoffman2014lsda, hoffman2015detector,tang2016large} from their paper. All models are trained using the \emph{same} backbone: VGG-CNN-F \cite{chatfield2014return} which is pretrained on the ImageNet classification dataset \cite{imagenet_cvpr09}. Also, similar to \cite{kumar2018dock}, we use the MCG \cite{pont2016multiscale} proposals instead of training the RPN. The models are evaluated using mAP at IoU threshold $0.5$ denoted as AP$_{50}$.

UniT beats the closest baseline, DOCK \cite{kumar2018dock}, by a significant margin (${\sim}16\%$ on AP$_{50}$), despite DOCK using more sophisticated similarity measures for knowledge transfer, which require additional data from VOC \cite{everingham2010pascal}, Visual Genome \cite{krishna2017visual}, and SUN \cite{xiao2010sun} datasets. As DOCK only transfers knowledge from \emph{base} class classifiers, this difference in performance can be attributed to UniT additionally effectively transferring knowledge from \emph{base} class regressors onto \emph{novel} class regressors (Eq. \ref{eq:reg}). It can also be noted that our work can be considered complimentary to DOCK, as we can easily integrate their richer similarity measures into our framework by modifying $\mathbf{S}(\mathbf{z})$ (Sec. \ref{sec:similarity}).

% We compare our approach to related methods in Table \ref{tab:semi-zero-shot}. Notably, DOCK \cite{kumar2018dock} sets the state-of-the-art and we borrowed performance of previous methods \cite{hoffman2014lsda, tang2016large, hoffman2015detector} from their paper. Similar to DOCK, we use the same backbone architecture: VGG-CNN-F \cite{chatfield2014return} which is pretrained on ImageNet classification dataset \cite{imagenet_cvpr09}, and used MCG \cite{arbelaez2014multiscale} object proposals.
% \RG{Justify improved performance: Our improved performance comes from the fact that we employ a better performing weakly-supervised method OICR \cite{tang2017multiple} as compared to WSDDN \cite{bilen2016weakly} used in DOCK. Also DOCK perform transfer on classifiers only, while we transfer box regressors too.}
% Please add the following required packages to your document preamble:
% \usepackage{booktabs}
\begin{table}[]
\centering
\scalebox{0.80}{
\begin{tabular}{@{}lcccc@{}}
\toprule
\multicolumn{1}{c}{Method}     & AP$_{50}$   & AP$_{S}$ & AP$_{M}$ & AP$_{L}$ \\ \midrule
LSDA \cite{hoffman2014lsda}                           & 4.6  & 1.2       & 5.1       & 7.8       \\
LSDA+Semantic \cite{tang2016large}                  & 4.7  & 1.1       & 5.1       & 8.0       \\
LSDA+MIL \cite{hoffman2015detector}                       & 5.9  & 1.5       & 8.3       & 10.7      \\
% Fine-tuned \cite{kumar2018dock}   & 10.8 & 1.2       & 8.9       & 18.6      \\
DOCK \cite{kumar2018dock}                           & 14.4 & 2.0       & 12.8      & 24.9      \\ \midrule
UniT (Ours)                           & \textbf{16.7}      & \textbf{3.2}           & \textbf{16.6}           & \textbf{27.3}           \\ \midrule
Full Supervision \cite{kumar2018dock} & 25.2 & 5.8       & 26.0      & 41.6      \\ \bottomrule
\end{tabular}
}
\vspace{-0.6em}
\caption{\textbf{Comparison to semi-supervised zero-shot.} All models are trained on VGG-CNN-F \cite{chatfield2014return} backbone.}
\label{tab:semi-zero-shot}
\vspace{-0.8em}
\end{table}
% Please add the following required packages to your document preamble:
% \usepackage{booktabs}
\begin{table}[]
\centering
% \setlength\tabcolsep{2.5pt}
% \resizebox{0.4\textwidth}{!}{
\scalebox{0.80}{
\begin{tabular}{@{}lccccc@{}}
\toprule
Method / Shots ($k$) & 12   & 33   & 55   & 76   & 96   \\ \midrule
NOTE-RCNN \cite{gao2019note}          & 14.1 & 14.2 & 17.1 & 19.8 & 19.9 \\
UniT (Ours)               & \textbf{14.7} & \textbf{17.4} & \textbf{19.3} & \textbf{20.9} & \textbf{22.1} \\ \bottomrule
\end{tabular}
}
\vspace{-0.6em}
\caption{\textbf{Comparison to semi-supervised few-shot.} All models are trained on Inception-ResNet-v2 \cite{szegedy2016inception} backbone. Mean Average Precision (mAP) on novel classes averaged over IoU thresholds in $[0.5: 0.05: 0.95]$ is reported.}
\vspace{-1.5em}
\label{tab:semi-few-shot}
\end{table}

% Please add the following required packages to your document preamble:
% \usepackage{booktabs}
% \usepackage{multirow}
% \usepackage{graphicx}
\begin{table*}[t]
\centering
\setlength\tabcolsep{2.5pt}
\resizebox{\textwidth}{!}{%
\begin{tabular}{@{}lccccccc@{\hskip 0.3in}cccccc@{\hskip 0.3in}cccccc@{}}
\toprule
                      & \multicolumn{1}{l}{} & \multicolumn{6}{c}{Novel Set 1}      & \multicolumn{6}{c}{Novel Set 2}      & \multicolumn{6}{c}{Novel Set 3}      \\ \midrule
Method / Shots  & \multicolumn{1}{l}{} & 0 & 1    & 2    & 3    & 5    & 10   & 0 & 1    & 2    & 3    & 5    & 10   & 0 & 1    & 2    & 3    & 5    & 10   \\ \midrule
Joint                 & FRCN \cite{Yan_2019_ICCV}      & - & 2.7  & 3.1  & 4.3  & 11.8 & 29.0 & - & 1.9  & 2.6  & 8.1  & 9.9  & 12.6 & - & 5.2  & 7.5  & 6.4  & 6.4  & 6.4  \\ \midrule
% Transfer              & FRCN \cite{Yan_2019_ICCV}         & - & 13.8 & 19.6 & 32.8 & 41.5 & 45.6 & - & 7.9  & 15.3 & 26.2 & 31.6 & 39.1 & - & 9.8  & 11.3 & 19.1 & 35.0 & 45.1
Transfer              & FRCN \cite{wang2020frustratingly}         & - & 15.2 & 20.3 & 29.0 & 40.1 & 45.5 & - & 13.4 & 20.6 & 28.6 & 32.4 & 38.8 & - & 19.6 & 20.8 & 28.7 & 42.2 & 42.1 \\ \midrule
\multirow{4}{*}{Few-Shot} & Kang \etal~\cite{Kang_2019_ICCV}         & - & 14.8 & 15.5 & 26.7 & 33.9 & 47.2 & - & 15.7 & 15.3 & 22.7 & 30.1 & 39.2 & - & 19.2 & 21.7 & 25.7 & 40.6 & 41.3 \\
                      & Wang \etal~\cite{Wang_2019_ICCV}              & - & 18.9 & 20.6 & 30.2 & 36.8 & 49.6 & - & 21.8 & 23.1 & 27.8 & 31.7 & 43.0 & - & 20.6 & 23.9 & 29.4 & 43.9 & 44.1 \\
                      & Yan \etal~\cite{Yan_2019_ICCV}           & - & 19.9 & 25.5 & 35.0 & 45.7 & 51.5 & - & 10.4 & 19.4 & 29.6 & 34.8 & 45.4 & - & 14.3 & 18.2 & 27.5 & 41.2 & 48.1 \\ 
                      & Wang \etal~\cite{wang2020frustratingly}           & - & 39.8 & 36.1 & 44.7 & 55.7 & 56.0 & - & 23.5 & 26.9 & 34.1 & 35.1 & 39.1 & - & 30.8 & 34.8 & 42.8 & 49.5 & 49.8 \\
                      \midrule
\multirow{1}{*}{Semi+Any Shot}       
% & UniT$_{budget}$ (Ours)                 &   &  &        &      &      &     &   &      &       &   &   &   &     &    &     &      &    &   \\     
& UniT (Ours)                 & \textbf{75.6}  &  \textbf{75.7}     & \textbf{75.8}       & \textbf{75.9}     & \textbf{76.1}      & \textbf{76.7}      & \textbf{56.9}  & \textbf{57.2}      & \textbf{57.4}      & \textbf{57.9}      & \textbf{58.2}    & \textbf{63.0}  & \textbf{67.5}      & \textbf{67.6}   & \textbf{68.1}      & \textbf{68.2}     & \textbf{68.6}      & \textbf{70.0        }       \\ \midrule
Fully-supervised      & FRCN         & \multicolumn{6}{c}{84.71}            & \multicolumn{6}{c}{82.89}            & \multicolumn{6}{c}{82.57}            \\ \bottomrule
\end{tabular}
}
\vspace{-0.15in}
\caption{{\bf Few-shot object detection on VOC.} FRCN $=$ Faster R-CNN with ResNet-101 backbone. Mean
AP$_{50}$ reported on {\em novel} classes; performance on {\em base} classes is reported in supplementary Section \ref{sec:baseclasses}.}
\vspace{-0.1in}
\label{tab:voc-few-shot}
\end{table*}
\begin{table}[]
\vspace{-0.15in}
\centering
\setlength\tabcolsep{2.5pt}
% \resizebox{0.5\textwidth}{!}{%
\scalebox{0.80}{
\begin{tabular}{@{}cccccccc@{}}
\toprule
% \#Shots                 & Method         & \multicolumn{3}{c}{Avg. Precision, IoU}                            & \multicolumn{3}{c}{Avg. Precision, Area}                           & \multicolumn{3}{c}{Avg. Recall, \#Dets}                            & \multicolumn{3}{c}{Avg. Recall, Area}                              \\ \midrule
             \#Shots           &                & AP             & AP$_{50}$                  & AP$_{75}$                 & AP$_S$                    & AP$_M$                    & AP$_L$                    \\ \midrule
$k=0$                   & UniT (Ours)           & 18.9                    & 36.1                     & 17.5                      & 8.7                     & 20.4                      & 27.6                       \\ \midrule
\multirow{6}{*}{$k=10$} & Transfer: FRCN \cite{Yan_2019_ICCV} & 6.5                  & 13.4                 & 5.9                  & 1.8                  & 5.3                  & 11.3                  \\
                        & Kang \etal~\cite{Kang_2019_ICCV}   & 5.6                  & 12.3                 & 4.6                  & 0.9                  & 3.5                  & 10.5                       \\
                        & Wang \etal~\cite{Wang_2019_ICCV}        & 7.1                  & 14.6                 & 6.1                  & 1.0                  & 4.1                  & 12.2                     \\
                        & Yan \etal~\cite{Yan_2019_ICCV}     & 8.7                  & 19.1                 & 6.6                  & 2.3                  & 7.7                  & 14.0                             \\
                        & Wang \etal~\cite{wang2020frustratingly} & 10.0 & - & 9.3 & - & - & -  \\
                        & UniT (Ours)           &  \textbf{21.7}                    & \textbf{40.8}                     & \textbf{20.6}                      & \textbf{9.1}                      & \textbf{23.8}                      & \textbf{31.3}                                           \\ \midrule
\end{tabular}%
}
\vspace{-0.15in}
\caption{\textbf{Few-shot object detection on COCO}. FRCN=Faster R-CNN with ResNet-50 backbone. Complete table is in supplementary Section \ref{sec:cocodet-full}.}
\vspace{-0.25in}
\label{tab:coco-few-shot}
\end{table}
\vspace{0.3em}
\noindent\textbf{Semi-supervised few-shot ($k\!>\!0$).} Table \ref{tab:semi-few-shot} compares the performance of our method with NOTE-RCNN \cite{gao2019note}, which is the only relevant baseline under this setting, on \emph{novel} classes. We follow the experimental setting described in \cite{gao2019note}, and our model is trained using the same backbone as NOTE-RCNN: Inception-Resnet-V2 \cite{szegedy2016inception} pretrained on the ImageNet classification dataset \cite{imagenet_cvpr09}, where the RPN is learned from the instance-level \emph{base} data. Similar to \cite{gao2019note}, we assume $k$ instance-level annotations for the \emph{novel} classes, where $k \in \{12, 33, 55, 76, 96\}$. To ensure fair comparison, the performance of NOTE-RCNN \cite{gao2019note} is taken from their published work\footnote{\cite{gao2019note} visualize their numbers as a plot instead of listing the raw values. As the authors were unreachable, Table \ref{tab:semi-few-shot} lists our best interpretation of the numbers shown in the plot.}. We report mAP on novel classes averaged over IoU thresholds in $[0.5: 0.05: 0.95]$.
% \Sid{Should we mention that for this experiment we finetuned the weak detectors as well?}      

UniT outperforms NOTE-RCNN \cite{gao2019note} on all values of $k$, providing an improvement of up to ${\sim}23\%$. Contrary to NOTE-RCNN that only trains \emph{novel} regressors on the $k$ shots, UniT benefits from effectively mapping information from \emph{base} regressors to \emph{novel} regressors. In addition, UniT also has the advantage of allowing end-to-end training while simultaneously being simple and interpretable. NOTE-RCNN, on the other hand, employs a complex multi-step bounding box mining framework that takes longer to train on \emph{novel} classes. 
Note that, in principle, one could incorporate the box mining mechanism into our framework as well.
% It should also be noted that one could, in principle, incorporate the NOTE-RCNN box mining mechanism into our framework as well.
% We compare with NOTE-RCNN \cite{gao2019note} which is the only method in this category in Table \ref{tab:semi-few-shot}. We use the same backbone Inception-Resnet-V2 \cite{szegedy2016inception} and the value of few-shot $k \in \{12, 33, 55, 76, 96\}$ as used in NOTE-RCNN.
% % \RG{Include \cite{uijlings2018revisiting}}
% \RG{Justify improved performance: Our improved performance can be explained in terms of simplicity of our approach ...}

% \textbf{Baselines.}
% We compare our approach to the state-of-the-art semi-supervised object detection methods: NOTE-RCNN \cite{gao2019note} and DOCK \cite{kumar2018dock}. We adopt the same backbone architecture for comparison, i.e., Inception-Resnet-V2 \cite{szegedy2016inception} in the case of NOTE-RCNN and VGG-CNN-F \cite{chatfield2014return} for DOCK. Both backbones are pretrained on ImageNet classification dataset \cite{imagenet_cvpr09}.

% \textbf{Results.}
% Table (ref) provides the results ... A key thing to note is that DOCK is similar to our weakly-supervised zero-shot setting where few-shot $k$ is $0$ for novel classes. On the other hand, NOTE-RCNN assumes the value of $k$ from the set $\{12, 33, 55, 76, 96\}$. \\ 
% - Discuss LSDA and Tang \etal as being discussed in NOTE-RCNN and DOCK. \\
% - Insert the use of mining in our approach as done in NOTE-RCNN \\
% - DOCK uses TTA \\

\vspace{-0.06in}
\subsection{Few-shot Object Detection and Segmentation}
\label{exp:vocexp}
\vspace{-0.06in}
\noindent\textbf{Datasets.}
We evaluate our models on VOC 2007 \cite{everingham2010pascal}, VOC 2012 \cite{everingham2015pascal}, and MSCOCO \cite{lin2014microsoft} as used in the previous few-shot object detection and segmentation works \cite{Kang_2019_ICCV, wang2020frustratingly, Wang_2019_ICCV, Yan_2019_ICCV}. For both detection and segmentation, we consistently follow the data splits introduced and used in \cite{Kang_2019_ICCV, Yan_2019_ICCV}. In case of VOC, we use VOC 07 test set ($5$k images) for evaluation and VOC 07+12 trainval sets ($16.5$k images) for training. The $20$ object classes are divided into $15$ base and $5$ novel classes with 3 different sets of class splits. For novel classes, we use images made available by Kang \etal~\cite{Kang_2019_ICCV} for $k$-shot fine-tuning. 
% where $k \in \{0, 1, 2, 3, 5, 10 \}$.
We report mean Average Precision (mAP) on novel classes and use a standard IoU threshold of $0.5$ \cite{everingham2010pascal}. Similarly, for the MSCOCO \cite{lin2014microsoft} dataset, consistent with \cite{Kang_2019_ICCV} we use $5$k images from the validation set for evaluation and the remaining $115$k trainval images for training. We assign 20 object classes from VOC as the novel classes and remaining 60 as the base classes. 
% The $k$-shots tasks are sampled as before with $k \in \{0, 10, 30 \}$, and 
We report the standard evaluation metric on COCO \cite{ren2015faster}. 

\vspace{0.2em}
\noindent\textbf{PASCAL VOC Detection.} Table \ref{tab:voc-few-shot} summarizes the results on VOC for three different novel class splits with different $k$-shot settings. Following \cite{wang2020frustratingly, Yan_2019_ICCV}, we assume Faster R-CNN \cite{ren2015faster} with ResNet-101 \cite{he2016deep} as the backbone which is pre-trained on ImageNet-1k \cite{russakovsky2015imagenet}. UniT outperforms the related state-of-the-art methods on all values of $k$, including the scenario where no instance-level supervision for novel classes is available ($k=0$), showing the effectiveness of transfer from base to novel classes. As UniT uses additional weak image-level data for \emph{novel} classes, this is not an equivalent comparison (see Sec. \ref{exp:annotation} comparisons under similar annotation budget). But we want to highlight that such data is readily available, much cheaper to obtain \cite{Bearman_2019_ECCV}, and leads to a significant improvement in performance. 
\begin{figure*}[t]
\captionsetup[subfigure]{labelformat=empty}
\begin{subfigure}{.2\textwidth}
\centering
\includegraphics[width=\textwidth, height=2.0cm]{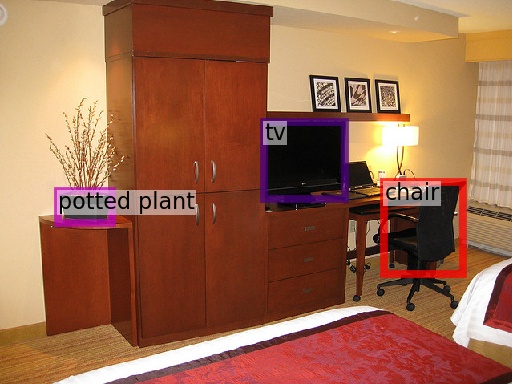}
\end{subfigure}%
\begin{subfigure}{.2\textwidth}
\centering
\includegraphics[width=\textwidth, height=2.0cm]{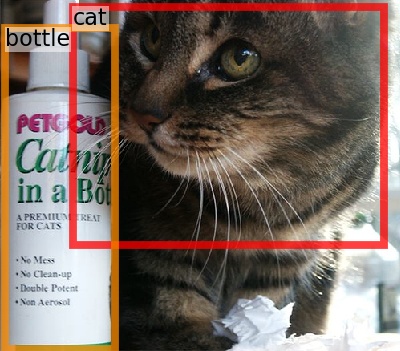}
\end{subfigure}%
\begin{subfigure}{.2\textwidth}
\centering
\includegraphics[width=\textwidth, height=2.0cm]{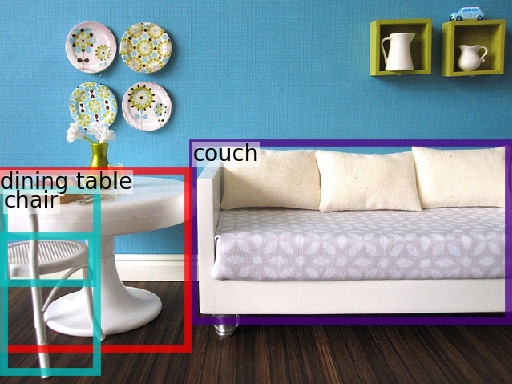}
\end{subfigure}%
\begin{subfigure}{.2\textwidth}
\centering
\includegraphics[width=\textwidth, height=2.0cm]{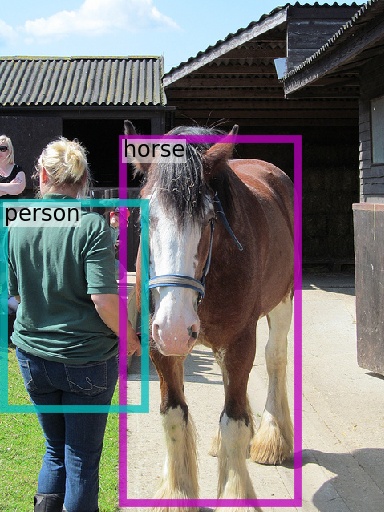}
\end{subfigure}%
\begin{subfigure}{.2\textwidth}
\centering
\includegraphics[width=\textwidth, height=2.0cm]{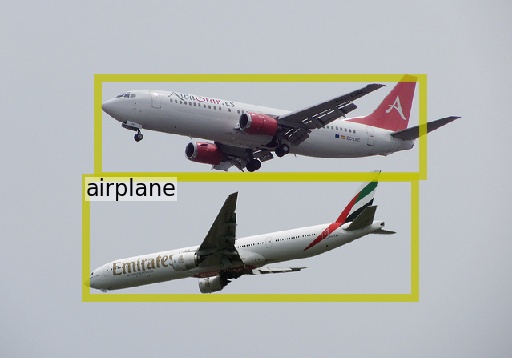}
\end{subfigure}
\begin{subfigure}{.2\textwidth}
\centering
\includegraphics[width=\textwidth, height=2.0cm]{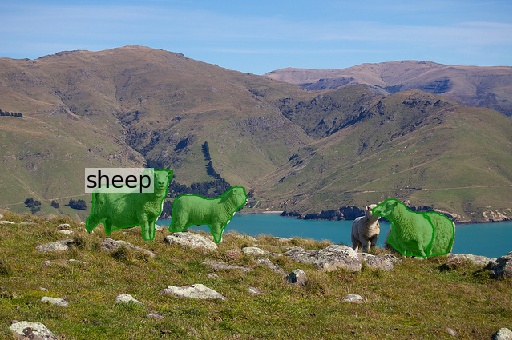}
\end{subfigure}%
\begin{subfigure}{.2\textwidth}
\centering
\includegraphics[width=\textwidth, height=2.0cm]{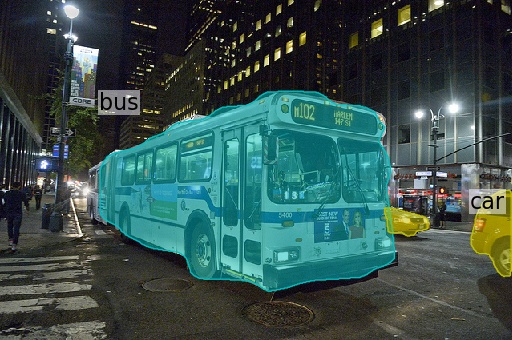}
\end{subfigure}%
\begin{subfigure}{.2\textwidth}
\centering
\includegraphics[width=\textwidth, height=2.0cm]{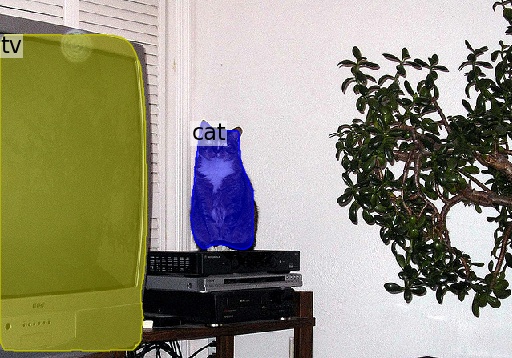}
\end{subfigure}%
\begin{subfigure}{.2\textwidth}
\centering
\includegraphics[width=\textwidth, height=2.0cm]{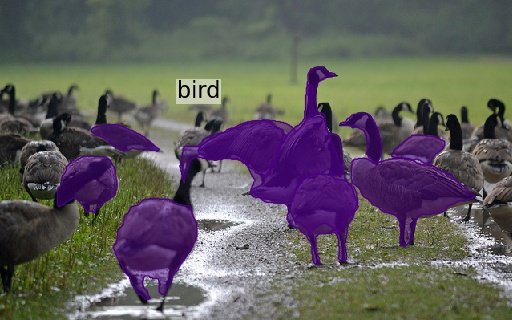}
\end{subfigure}%
\begin{subfigure}{.2\textwidth}
\centering
\includegraphics[width=\textwidth, height=2.0cm]{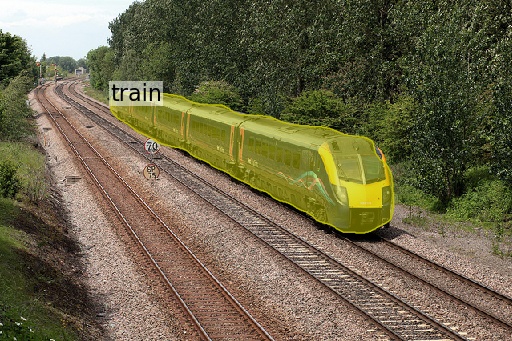}
\end{subfigure}
\vspace{-0.06in}
\caption{\textbf{Qualitative Visualizations.} Semi-supervised zero-shot ($k=0$) detection (top) and instance segmentation (bottom) performance on \emph{novel} classes in MS-COCO (%best viewed with magnification; for each image, one 
color $=$ object category). Additional visualizations in Section \ref{sec:additionalviz} of the appendix.}
\label{fig:cocoviz}
\vspace{-0.2in}
\end{figure*}

\vspace{0.3em}
\noindent\textbf{MS-COCO Detection.} 
Table \ref{tab:coco-few-shot} describes the results on COCO dataset. Similar to \cite{Yan_2019_ICCV, wang2020frustratingly}, we use ImageNet \cite{imagenet_cvpr09} pretrained ResNet-50 \cite{he2016deep} as the backbone. We observe similar trends as above. In addition, we note that our performance consistently increases with the value of $k$ even on larger datasets, showing that UniT is effective and flexible in scaling with the degree of instance-level supervision ranging from zero to a few. The full table is in appendix Section \ref{sec:cocodet-full}. Figure \ref{fig:cocoviz} shows qualitative results, indicating our method is able to correctly detect \emph{novel} classes.
% In addition, we observe larger margin of improvement with more classes present in COCO.

\vspace{0.3em}
\noindent\textbf{MS-COCO Segmentation.} Table \ref{tab:coco-few-segm} summarizes the results. Similar to \cite{Yan_2019_ICCV}, we choose an ImageNet\cite{imagenet_cvpr09} pretrained ResNet-50 \cite{he2016deep} backbone. UniT consistently improves over \cite{Yan_2019_ICCV}, demonstrating that our approach is not limited to bounding boxes, and is able to generalize over the type of downstream structured label by effectively transferring information from \emph{base} segmentations to \emph{novel} segmentations. The full table is provided in appendix Section \ref{sec:cocoseg}. Figure \ref{fig:cocoviz} shows some qualitative results on $k=0$ for \emph{novel} classes.
\begin{table}[]
\vspace{-0.15in}
\centering
\setlength\tabcolsep{2.5pt}
\scalebox{0.80}{
\begin{tabular}{@{}cc@{\hskip 0.15in}c@{\hskip 0.15in}cccccc@{}}
\toprule
\#Shots           & Method            &      & AP & AP$_{50}$ & AP$_{75}$ & AP$_{S}$ & AP$_M$ & AP$_L$ \\
\midrule
\multirow{2}{*}{$k=0$} & \multirow{2}{*}{UniT (Ours)} & Box  & 20.2	& 36.8 &	19.5 & 8.5 & 20.9 & 28.9     \\
                  &                   & Mask & 17.6 & 32.7 & 17.0 & 5.6 & 17.6 & 27.7 \\
\midrule
\multirow{4}{*}{$k=10$} & \multirow{2}{*}{Yan \etal~\cite{Yan_2019_ICCV}} & Box & 5.6 & 14.2 & 3.0  & 2.0 & 6.6 & 8.8 \\
& & Mask & 4.4 & 10.6 & 3.3  & 0.5 & 3.6 & 7.2\\   
    \addlinespace[0.1cm]
    \cdashline{2-9}
    \addlinespace[0.1cm]
    & \multirow{2}{*}{UniT (Ours)} & Box  & \textbf{22.8}&	\textbf{41.6}&	\textbf{21.9}&	\textbf{9.4}&	\textbf{24.4}&	\textbf{32.3}    \\

                  &                   & Mask & \textbf{20.5}&	\textbf{38.6}&	\textbf{19.7}&	\textbf{6.0}&	\textbf{20.5}&	\textbf{31.8} \\ \midrule  
\end{tabular}
}
\vspace{-0.8em}
\caption{\textbf{Few-shot instance segmentation on COCO}. Complete table is in supplementary Section \ref{sec:cocoseg}.}
\label{tab:coco-few-segm}
\vspace{-1.8em}
\end{table}

\vspace{0.3em}
\noindent\textbf{Ablation.} A complete ablation study on MSCOCO \cite{lin2014microsoft} is provided in appendix Section \ref{sec:ablation}.
We report performance on the novel split used by \cite{Yan_2019_ICCV}, starting with only weak detectors and progressively adding the terms in Eq.(\ref{eq:cls-base}), (\ref{eq:cls}), (\ref{eq:reg}), and (\ref{eq:seg}). 
% We add Top-K transfer from base classes (delta base term) which is the criteria based on LSDA \cite{hoffman2014lsda}. 
% Top-K transfer from base classes (instance-level transfer) is used as a baseline (akin to LSDA \cite{hoffman2014lsda}).
Weighting with visual and lingual similarity results in $+1.4$ AP$_{50}$ improvement (Eq. (\ref{eq:cls})), transfer from \emph{base} regressors (Eq. (\ref{eq:reg})) provides an additional $+7$ AP$_{50}$ imrovement. Finally, transfer from \emph{base} class segmentations (Eq. (\ref{eq:seg})) leads to an added gain of $+7.5$ on mask AP$_{50}$.
% We then use our similarity matrix to weigh the deltas and found further \RG{XX\%} 
% improvement; combination with visual is an additional $0.9$ mAP.
% . In particular, we also ablate over lingual and visual component of our similarity matrix. 
% Finally, transfer from base classes for regressors leads to added gain of $12.9$. % mAP.
\vspace{-0.04in}
\subsection{Limited Annotation Budget}
\label{exp:annotation}
\vspace{-0.08in}
Compared to approaches in the few-shot detection (and segmentation) domain like \cite{Kang_2019_ICCV,wang2020frustratingly,Wang_2019_ICCV,Yan_2019_ICCV}, UniT assumes additional image-level annotations for \emph{novel} classes. We argue this is a reasonable assumption considering that such annotations are readily available in abundance for thousands of object classes (${\sim}22$K in ImageNet \cite{imagenet_cvpr09} and ${\sim}20$K in Open Image v4 dataset \cite{kuznetsova2020open}). Experiments in Section \ref{exp:vocexp} further highlight the performance improvements possible by using such inexpensive data. However, this raises an interesting question as to what form of supervision could be more valuable, if one is to collect it. To experiment with this, we conceptually impose an annotation budget that limits the number of \emph{novel} class image-level annotations our approach can use. For object detection on VOC \cite{everingham2015pascal}, we assume $7$ image-level annotations can be generated in the same time as $1$ instance-level annotation. The $7$ conversion factor between object instance labels and weakly-supervised image-level labels is motivated by the timings reported in \cite{Bearman_2019_ECCV} and is a \emph{conservative} estimate (additional details in appendix Section \ref{sec:budget-details})\footnote{This factor is expected to be higher in practice, as we don't consider situations where boxes/masks are rejected and need to be redrawn \cite{papadopoulos2017training}.}. Therefore, for each value of $k$ in a few-shot setup, we train a variant of UniT that assumes only $7\times k$ image-level annotations for \emph{novel} classes, which is referred to as UniT$_{budget=k}$. We then compare the \emph{zero-shot} performance of UniT$_{budget=k}$ against the corresponding $k$-shot generalized object detection benchmarks\footnote{These benchmarks use multiple random splits as opposed to curated splits used in \cite{Kang_2019_ICCV} and Table \ref{tab:voc-few-shot}. As per \cite{wang2020frustratingly}, this helps reduce variance.} reported in \cite{wang2020frustratingly}. This setting allows for an apples-to-apples comparison with the baselines, while simultaneously enabling the analysis of the relative importance of image-level annotations when compared to instance-level annotations.

Please refer to Section \ref{exp:vocexp} for details on the dataset and setup. Table \ref{tab:anno-voc} summarizes the results on VOC for three different novel class splits with different $k$-shot settings. For accurate results, on each split and $k$-shot, we perform 10 repeated runs of UniT$_{budget=k}$, selecting a random set of $7 \times k$ weakly-labelled \emph{novel} class images each time. Following \cite{wang2020frustratingly}, we assume ResNet-101 \cite{he2016deep} as the backbone. 
% which is pre-trained on ImageNet-1k \cite{russakovsky2015imagenet}. 
In table \ref{tab:annotation-ft}, for \emph{novel} split 1, we further analyse the relative importance of image-level to instance-level annotations. For a fixed budget equivalent to $10$ instance-level annotations, we experiment with using different proportions of image and instance-level annotations, and report mean AP$_{50}$ on \emph{novel} classes computed across $10$ repeated runs.

Even under the constraint of equal budget, UniT$_{budget=k}$ outperforms the state-of-the-art \cite{wang2020frustratingly} on multiple splits. This highlights three key observations: i) weak image-level supervision, which is cheaper to obtain \cite{Bearman_2019_ECCV}, provides a greater `bang-for-the-buck' when compared to instance-level supervision, ii) our structured transfer from \emph{base} classes is effective even when the amount of \emph{novel} class supervision is limited, and iii) from Table \ref{tab:annotation-ft}, in a low-shot and fixed budget setting, it is more beneficial to just use weak supervision (instead of some combination of both). Furthermore, as our approach is agnostic to the type of weak detector used, employing better weak detectors like \cite{tang2018pcl, arun2019dissimilarity} could further improve the performance of UniT$_{budget=k}$.
\begin{table}[t]
\centering
\resizebox{0.45\textwidth}{!}{%
\begin{tabular}{@{}ccccc@{}}
\toprule
\#Shots              & Method                                  & Split 1        & Split 2        & Split 3        \\ \midrule
\multirow{3}{*}{1}  & Kang \etal~\cite{Kang_2019_ICCV}        & $14.2 \pm 1.7$ & $12.3 \pm 1.9$ & $12.5 \pm 1.6$ \\
                    & Wang \etal~\cite{wang2020frustratingly} & $25.3 \pm 2.2$ & $\mathbf{18.3 \pm 2.4}$ & $17.9 \pm 2.0$ \\
                    & UniT$_{budget=1}$ (Ours)                  &  $\mathbf{28.3 \pm 2.0}$              & $17.0 \pm 1.9$               & $\mathbf{26.2 \pm 2.5}$               \\ \midrule
\multirow{3}{*}{5}  & Kang \etal~\cite{Kang_2019_ICCV}        & $36.5 \pm 1.4$ & $31.4 \pm 1.5$ & $33.8 \pm 1.4$ \\
                    & Wang \etal~\cite{wang2020frustratingly} & $47.9 \pm 1.2$ & $34.1 \pm 1.4$ & $40.8 \pm 1.4$ \\
                    & UniT$_{budget=5}$ (Ours)                  &     $\mathbf{50.9 \pm 1.4}$           &  $\mathbf{36.2 \pm 1.7}$              &  $\mathbf{47.4 \pm 1.2}$               \\ \midrule
\multirow{2}{*}{10} & Wang \etal~\cite{wang2020frustratingly} & $52.8 \pm 1.0$ & $39.5 \pm 1.1$ & $45.6 \pm 1.1$ \\
                    & UniT$_{budget=10}$ (Ours)                  &      $\mathbf{59.0 \pm 1.5}$          & $\mathbf{40.8 \pm 1.3}$                & $\mathbf{52.9 \pm 1.1}$                \\ \bottomrule
\end{tabular}
}
\vspace{-0.5em}
\caption{\textbf{Limited annotation budget.} Averaged AP$_{50}$ for $10$ random runs with 95\% confidence interval estimate
\cite{wang2020frustratingly}.}
\label{tab:anno-voc}
\vspace{-1.2em}
\end{table}
\begin{table}[]
\centering
\scalebox{0.85}{
\begin{tabular}{@{}c@{\hspace{0.5\tabcolsep}}c@{\hspace{0.5\tabcolsep}}cc@{}}
\toprule
Method                                  & \begin{tabular}[c]{@{}c@{}}Weak \\ Anno.(\%)\end{tabular} & \begin{tabular}[c]{@{}c@{}}Instance \\ Anno.(\%)\end{tabular} & AP$_{50}$      \\ \midrule
Wang \etal~\cite{wang2020frustratingly} + $10$-Shots & 0                                                        & 100                                                      & $52.8 \pm 1.0$ \\
UniT$_{budget=1}$ + $9$-Shots           & 10                                                       & 90                                                       & $49.2 \pm 0.6$ \\
UniT$_{budget=5}$ + $5$-Shots           & 50                                                       & 50                                                       & $54.0 \pm 0.8$ \\
UniT$_{budget=10}$ + $0$-Shots          & 100                                                      & 0                                                        & $\mathbf{59.0 \pm 1.5}$ \\ \bottomrule
\end{tabular}
}
\vspace{-0.5em}
\caption{\textbf{Using different annotation proportions}. For the same budget, we vary the amount of image/instance level annotation. Averaged AP$_{50}$ for $10$ random runs with 95\% confidence interval estimate of the mean values \cite{wang2020frustratingly} is shown.}
\label{tab:annotation-ft}
\vspace{-1.7em}
\end{table}

\vspace{-0.08in}
\section{Discussion and Conclusion}
\vspace{-0.1in}
% {\bf Discussion and Conclusion.}
We propose an intuitive semi-supervised model that is applicable to a range of supervision: from zero to a few instance-level samples per {\em novel} class.
For {\em base} classes, our model learns a mapping from weakly-supervised to fully-supervised detectors/segmentors. By leveraging 
%  learning and leveraging visual and lingual 
similarities between the {\em novel} and {\em base} classes, we transfer those mappings to obtain detectors/segmentors for {\em novel} classes; refining them with a few novel class instance-level annotated samples, if available. This versatile paradigm works significantly better than traditional semi supervised and few-shot detection and segmentation methods. 
%  We observe greater improvements when number of classes is large (COCO). 

%- Comment on COCO v/s VOC results, and the fact that COCO has more number of classes and exhibits greater potential for an effective transfer. \\
%- Not limited to the type of weak supervision. Could be interesting to see points, scribbles, etc.

\begin{ack}
This work was funded, in part, by the Vector Institute for AI, Canada CIFAR AI Chair, NSERC CRC and an NSERC DG and Discovery Accelerator Grants.

This material is based upon work supported by the United States Air Force Research Laboratory (AFRL) under the Defense Advanced Research Projects Agency (DARPA) Learning with Less Labels (LwLL) program (Contract No.FA8750-19-C-0515). The views and conclusions contained herein are those of the authors and should not be interpreted as necessarily representing the official policies or endorsements, either expressed or implied, of DARPA or the U.S. Government.

Resources used in preparing this research were provided, in part, by the Province of Ontario, the Government of Canada through CIFAR, and companies sponsoring the Vector Institute \url{www.vectorinstitute.ai/\#partners}. 

Additional hardware support was provided by John R. Evans Leaders Fund CFI grant and Compute Canada under the Resource Allocation Competition award. 

Finally, we would like to sincerely thank Bicheng Xu, Issam Laradji and Prof. Frank Wood for valuable feedback and discussions. 

% Use unnumbered first level headings for the acknowledgments. All acknowledgments go at the end of the paper before the list of references. Moreover, you are required to declare funding (financial activities supporting the submitted work) and competing interests (related financial activities outside the submitted work). More information about this disclosure can be found at: \url{https://neurips.cc/Conferences/2020/PaperInformation/FundingDisclosure}.

% Do {\bf not} include this section in the anonymized submission, only in the final paper. You can use the \texttt{ack} environment provided in the style file to autmoatically hide this section in the anonymized submission.
\end{ack}

{\small
\bibliographystyle{ieee_fullname}
\bibliography{egbib}
}
\clearpage
% {\small
% \bibliographystyleFig{alpha}
% \bibliography{egbib}
% }

\clearpage
% \title{Supplementary Material: Weakly-supervised Any-shot Object Detection}
% \maketitle
% \setcounter{page}{1}
% \setcounter{section}{0}
% \setcounter{table}{0}
% \setcounter{figure}{0}
% \setcounter{equation}{0}

% \resetlinenumber
% \twocolumn[
% \begin{center}
%     {\LARGE \textbf{Supplementary for \#4174\\ UniT: Unified Knowledge Transfer for Any-shot Object Detection and Segmentation\\}}
% \end{center}
% \vspace{1em}
% ]
{\noindent\LARGE \textbf{Appendix}}
\appendix
\setcounter{table}{0}
\setcounter{figure}{0}
\setcounter{equation}{0}
\renewcommand{\thetable}{A\arabic{table}}
\renewcommand{\thefigure}{A\arabic{figure}}
% \newcount\cvprrulercount
% Throughout this supplementary, [\textcolor{green}{x}] refers to citations from the main paper, and [\textcolor{Magenta}{x}] refers to additional citations which are provided at the end of \textbf{this document} (on page 11). 
% For ease of reference, the table/figure numbers used in this supplementary are carried over from the main paper.
\vspace{-0.05in}
\section{Semi-Supervised Any-shot Segmentation}
\label{sec:segmentation}
\vspace{-0.05in}
% PARAMS LIST FOR FAST R-CNN AND MASK-HEAD
% (box_predictor): FastRCNNOutputLayers(
%   (cls_score): Linear(in_features=2048, out_features=81, bias=True)
%   (bbox_pred): Linear(in_features=2048, out_features=320, bias=True)
% )
% (mask_head): MaskRCNNConvUpsampleHead(
%   (deconv): ConvTranspose2d(2048, 256, kernel_size=(2, 2), stride=(2, 2))
%   (predictor): Conv2d(256, 80, kernel_size=(1, 1), stride=(1, 1))
% )

% deconv: (2048, 14, 14) --> (256, 28, 28)
% predictor: (256, 28, 28) --> (80, 28, 28)

% Citation for Transposed Conv (needed?)
% @article{dumoulin2016guide,
%   title={A guide to convolution arithmetic for deep learning},
%   author={Dumoulin, Vincent and Visin, Francesco},
%   journal={arXiv preprint arXiv:1603.07285},
%   year={2016}
% }
In Equation \ref{eq:seg} of the main paper we discuss the formulation of the \emph{novel} segmentation head $f_{\mathbf{\W}_{novel}^{seg}}(\mathbf{z})$, which is obtained via a structured transfer from the \emph{base} segmentation head $f_{\mathbf{\W}_{base}^{seg}}(\mathbf{z})$. We provide more details on how this is implemented in practice.
Our implementation of segmentation module can be seen an extension to the Fast R-CNN \cite{girshick2015fast} pipeline described in Section~\ref{sec:approach} of the main paper. In particular, the segmentation module consists of a transposed-convolution layer ({\tt nn.ConvTranspose2D}), followed by {\tt ReLU}, and a $1 \times 1$ convolution layer ({\tt nn.Conv2D}). The feature vector $\mathbf{z}_{i,j}$ for a proposal $j$ in image $i$ is of dimension $(2048 \times 7 \times 7)$ where $2048$ is the number of channels and $7$ is the spatial resolution of the proposal's feature map. The segmentation module upsamples $\mathbf{z}$ (as in the main paper we drop $i,j$ indexing) using the transposed convolution layer with a kernel size of $2$, and then produces a class-specific mask using a $1 \times 1$ convolution layer. The resulting mask output is of size $(|\mathcal{C}| \times 14 \times 14)$, where $\mathcal{C}$ is the total number of  object classes. 

In Eq. \ref{eq:seg} of the main paper, $f_{\mathbf{\W}_{*}^{seg}}(\cdot)$ is the class-specific output of the segmentation module obtained after the $1 \times 1$ convolution. During training, we use the same loss formulation for $\mathcal{L}_{mask}$ as described in \cite{he2017mask}, where a per-pixel binary cross-entropy loss is used. During inference, the mask is interpolated to fit the regressed proposal (as obtained by Eq.(\ref{eq:reg-base}) and Eq.(\ref{eq:reg}) in the main paper) to produce the final output. For the semi-supervised zero-shot $(k=0)$ scenario, the predictions for \emph{novel} classes can be done as in Eq. (\ref{eq:seg}) but omitting the ``instance-level direct adaptation” term.

\vspace{-0.05in}
\section{Implementation Details}
\label{sec:implementation}
\vspace{-0.05in}
% Restate that we will release the code and pre-trained models\\
% Talk about RCNN loss (Eq 9) \\
For base-training, we train our model jointly with weak-supervision and base-detection/-segmentation losses with equal weighting (see Section \ref{sec:loss}). In particular, we use image-level data for all the classes to train the weakly-supervised OICR \cite{tang2017multiple} branch, and use detection/segmentation data of base classes for training base detectors/segmentors. Unless pretrained proposals are used, the proposals used for training weakly-supervised branch come from the RPN trained using the base-detection branch. For the zero-shot experiments ($k=0$) in Section \ref{exp:semisup}, similar to the baselines, we replace the RPN with the precomputed MCG proposals \cite{pont2016multiscale}. 
% We employ task-based training procedure where each task consist of $5$ support images per class and $8$ query images in a batch for base-training. 
We use $4$ Nvidia Tesla T4 GPUs to train models. We build on top of Detectron2 \cite{wu2019detectron2} library written in PyTorch \cite{paszke2019pytorch} framework, and unless mentioned, we keep their default configuration: SGD as the optimizer, RoI Align \cite{he2017mask} as the pooling method, ResNet layer sizes/parameters. We use the standard loss for Faster R-CNN, {\em i.e.}, cross-entropy loss for classifier and smooth-L1 loss for regressor as described in \cite{ren2015faster}.

Note that, in the following text, an iteration refers to a gradient step, and \emph{not} the total number of examples in the training set.

% Architecture details:
% - RoIAlign type (compare it with what Meta R-CNN uses)
\vspace{-0.5em}
\paragraph{Semi-Supervised Object Detection}
We train on the MSCOCO 2015 \cite{lin2014microsoft} data for semi-supervised zero-shot ($k=0$) and MSCOCO 2017 \cite{lin2014microsoft} for semi-supervised few-shot ($k>0$) experiments. We use $270$K iterations (default in Detectron2 \cite{wu2019detectron2}) to account for more data. For fine-tuning, we use $1000$ iterations for $12$-shot, $3000$ iterations for $33$-shot, $6000$ iterations for $55$-shot, $8000$ iterations for $76$-shot, and $10000$ iterations for $96$-shot experiments. 

\vspace{-0.5em}
\paragraph{Few-shot Object Detection: VOC.}

We train on VOC 07 + 12 dataset. We use a learning rate of $0.02$ over $30$K iterations. We decrease the learning rate by a factor of $10$ at $12$K and $24$K iteration. 
% (\textcolor{red}{Do we wanna mention how we select our model?})

For fine-tuning, we are given $k$-shot data for novel classes where $k \in \{ 1, 2, 3, 5, 10\}$. We linearly scale the number of SGD steps for optimizing over the $k$-shot data. In particular, we choose $50$ iterations for $k=1$, $100$ iterations for $k=2$, and similarly linearly scale to $500$ iterations for $k=10$.
\vspace{-0.5em}

\paragraph{Few-shot Object Detection: COCO.}
In the case of COCO dataset, we use $270$K iterations (default in Detectron2 \cite{wu2019detectron2}) to account for more data as compared to VOC. For fine-tuning, we use $500$ iterations for $10$-shot and $1500$ iterations for $30$-shot experiment.

% Base training (VOC 07 + 12) \\
% - LR: 0.02 \\
% - 30K iterations \\
% - schedule (12K, 24K) \\
% - BS: 8 (2 images per GPU) \\
% - Optimizer (SGD) \\
% Detectron2, Selecting model based on val performance \\

% Fine-tuning () \\ 
% - Linearly scale the iterations with $k$ \\
% - COCO iterations: 10 - 500, 30 - 1500 \\

% Weakly-sup \\
% - VGG16 \\
% - LR: 0.005 \\
% - BS: 8 (2 images per GPU) \\
% - 40K iterations (30K, )\\

% COCO: default \\
% - 270K (210, 240K) \\
% - BS: 8 (2 images per GPU) \\
% As mentioned in the main paper, {\bf we will release the code and pre-trained models} shortly. 

\paragraph{Weakly-supervised Object Detection}
The results are shown in Section \ref{exp:weaklyexp}. Here we use a pre-trained VGG-16 \cite{simonyan2014very} as the backbone to be consistent with the prior state-of-the-art works \cite{bilen2016weakly,tang2017multiple,arun2019dissimilarity,ren2020instance}. We use a learning rate of $0.001$ over $40$K iterations for optimization, dropping it down to $0.0001$ for the next $30$K iterations.

\vspace{0.5em}
\noindent
We will also make all our code \textbf{publicly available.}

\begin{table*}[]
\centering
\resizebox{\textwidth}{!}{%
\begin{tabular}{@{}lc@{\hskip 0.5in}cccccc@{\hskip 0.5in}cccccc@{}}
\toprule
% \#Shots                 & Method         & \multicolumn{3}{c}{Avg. Precision, IoU}                            & \multicolumn{3}{c}{Avg. Precision, Area}                           & \multicolumn{3}{c}{Avg. Recall, \#Dets}                            & \multicolumn{3}{c}{Avg. Recall, Area}                              \\ \midrule
             \#Shots           &                & AP             & AP$_{50}$                  & AP$_{75}$                 & AP$_S$                    & AP$_M$                    & AP$_L$                    & AR$_1$                    & AR$_{10}$                   & AR$_{100}$                  & AR$_{S}$                    & AR$_{M}$                    & AR$_L$                    \\ \midrule
$k=0$                   & UniT (Ours)           & 18.9                    & 36.1                     & 17.5                      & 8.7                      & 20.4                      & 27.6                      & 19.1                      & 33.3                     & 35.0                     & 16.5                      & 35.5                      & 48.5                     \\ \midrule
\multirow{6}{*}{$k=10$} & Transfer: FRCN \cite{Yan_2019_ICCV} & 6.5                  & 13.4                 & 5.9                  & 1.8                  & 5.3                  & 11.3                 & 12.6                 & 17.7                 & 17.8                 & 6.5                  & 14.4                 & 28.6                 \\
                        & Kang \etal~\cite{Kang_2019_ICCV}   & 5.6                  & 12.3                 & 4.6                  & 0.9                  & 3.5                  & 10.5                 & 10.1                 & 14.3                 & 14.4                 & 1.5                  & 8.4                  & 28.2                 \\
                        & Wang \etal~\cite{Wang_2019_ICCV}        & 7.1                  & 14.6                 & 6.1                  & 1.0                  & 4.1                  & 12.2                 & 11.9                 & 15.1                 & 15.5                 & 1.7                  & 9.7                  & 30.1                 \\
                        & Yan \etal~\cite{Yan_2019_ICCV}     & 8.7                  & 19.1                 & 6.6                  & 2.3                  & 7.7                  & 14.0                 & 12.6                 & 17.8                 & 17.9                 & 7.8                  & 15.6                 & 27.2                 \\
                        & Wang \etal~\cite{wang2020frustratingly} & 10.0 & - & 9.3 & - & - & - & - & - & -& - & - & - \\
                        & UniT (Ours)           &  \textbf{21.7}                    & \textbf{40.8}                     & \textbf{20.6}                      & \textbf{9.1}                      & \textbf{23.8}                      & \textbf{31.3}                      & \textbf{21.1}                     & \textbf{35.1}                      & \textbf{36.4}                      & \textbf{16.5}                     & \textbf{37.5}                      & \textbf{51.0}                     \\ \midrule
\multirow{6}{*}{$k=30$} & Transfer: FRCN \cite{Yan_2019_ICCV} & 11.1                 & 21.6                 & 10.3                 & 2.9                  & 8.8                  & 18.9                 & 15.0                 & 21.1                 & 21.3                 & 10.1                 & 17.9                 & 33.2                 \\
                        & Kang \etal~\cite{Kang_2019_ICCV}   & 9.1                  & 19.0                 & 7.6                  & 0.8                  & 4.9                  & 16.8                 & 13.2                 & 17.7                 & 17.8                 & 1.5                  & 10.4                 & 33.5                 \\
                        & Wang \etal~\cite{Wang_2019_ICCV}        & 11.3                 & 21.7                 & 8.1                  & 1.1                  & 6.2                  & 17.3                 & 14.5                 & 18.9                 & 19.2                 & 1.8                  & 11.1                 & 34.4                 \\
                        & Yan\etal~\cite{Yan_2019_ICCV}     & 12.4                 & 25.3                 & 10.8                 & 2.8                  & 11.6                 & 19.0                 & 15.0                 & 21.4                 & 21.7                 & 8.6                  & 20.0                 & 32.1                 \\
                        & Wang \etal~\cite{wang2020frustratingly} & 13.7 & - & 13.4 & - & - & - & - & - & -& - & - & - \\
                        & UniT (Ours)           &  \textbf{23.1}                    & \textbf{43.0}                     & \textbf{21.6}                      & \textbf{9.8}                      & \textbf{25.3}                      & \textbf{33.8}                      & \textbf{22.4}                     & \textbf{36.7}                     & \textbf{37.9}                     & \textbf{16.5}                      & \textbf{38.7}                      & \textbf{53.3}                      \\  \bottomrule
\end{tabular}%
}
\vspace{-0.15in}
\caption{\textbf{Complete few-shot object detection results on COCO}. FRCN=Faster R-CNN with ResNet-50 \cite{he2016deep} backbone. Similar to \cite{wang2020frustratingly, Yan_2019_ICCV}, we use ResNet-50 as the backbone.}
% \vspace{-0.20in}
\label{tab:coco-det-full}
\end{table*}
\section{Few-shot Object Detection on MSCOCO}
\label{sec:cocodet-full}
% \vspace{-2.5em}
As described in Section \ref{exp:vocexp} of the main paper, we compare the performance of UniT against state-of-the-art approaches on the task of Few-Shot Object Detection on the MSCOCO dataset \cite{lin2014microsoft}. Please refer to Section \ref{exp:vocexp} for task and dataset specifications. Similar to \cite{wang2020frustratingly, Yan_2019_ICCV}, we choose ResNet-50 \cite{he2016deep} as the backbone. The k-shot tasks are sampled following \cite{Yan_2019_ICCV} with $k \in \{0, 10, 30\}$. Due to lack of space, results in Table \ref{tab:coco-few-shot} of the main paper only compared our model against the baselines on $10$-shots. Here we present the complete comparison in Table \ref{tab:coco-det-full}.

As can be seen from Table~\ref{tab:coco-det-full} we get a significant boost in performance even on large number of shots $k=30$. As UniT uses additional weak image-level data for \emph{novel} classes, this is not an equivalent comparison (see Sec. \ref{exp:annotation} of the main paper for comparisons under similar annotation budget). But we want to highlight that such data is readily available, much cheaper to obtain \cite{Bearman_2019_ECCV}, and leads to a significant improvement in performance.
% Please add the following required packages to your document preamble:
% \usepackage{booktabs}
% \usepackage{multirow}
% \usepackage{graphicx}
\begin{table*}[t]
% \vspace{-0.06in}
\centering
\setlength\tabcolsep{2.5pt}
% \resizebox{\textwidth}{!}{%
\scalebox{0.9}{
\begin{tabular}{@{}cc@{\hskip 0.3in}cccccc@{\hskip 0.3in}cccccc@{}}
\toprule
                    &            & \multicolumn{6}{c}{Box}             & \multicolumn{6}{c}{Mask}            \\ \midrule
\#Shots                & Method     & AP  & AP$_{50}$ & AP$_{75}$ & AP$_{S}$ & AP$_M$ & AP$_L$ & AP  & AP$_{50}$ & AP$_{75}$ & AP$_S$ & AP$_M$ & AP$_L$ \\ \midrule
$k=0$                   & Ours       & 20.2	& 36.8	& 19.5 &	8.5 &	20.9 & 28.9&		17.6	&32.7	&17.0	&5.6	&17.6	&27.7 \\ \midrule
\multirow{2}{*}{$k=5$}  & Yan \etal~\cite{Yan_2019_ICCV} & 3.5 & 9.9  & 1.2  & 1.2 & 3.9 & 5.8 & 2.8 & 6.9  & 1.7  & 0.3 & 2.3 & 4.7 \\
                    & Ours       &     \textbf{22.1}&	\textbf{39.9}&	\textbf{21.7}&	\textbf{9.2}&	\textbf{23.0}&	\textbf{31.7}& \textbf{20.0}&	\textbf{37.5}&	\textbf{19.0}&	\textbf{6.1}&	\textbf{19.4}&	\textbf{31.1}      \\ \midrule
\multirow{2}{*}{$k=10$} & Yan \etal~\cite{Yan_2019_ICCV} & 5.6 & 14.2 & 3.0  & 2.0 & 6.6 & 8.8 & 4.4 & 10.6 & 3.3  & 0.5 & 3.6 & 7.2 \\
                    & Ours       &     \textbf{22.8}&	\textbf{41.6}&	\textbf{21.9}&	\textbf{9.4}&	\textbf{24.4}&	\textbf{32.3}& \textbf{20.5}&	\textbf{38.6}&	\textbf{19.7}&	\textbf{6.0}&	\textbf{20.5}&	\textbf{31.8}    \\ 
                    \midrule
\multirow{2}{*}{$k=20$} & Yan \etal~\cite{Yan_2019_ICCV} & 6.2 & 16.6 & 2.5  & 1.7 & 6.7 & 9.6 & 6.4 & 14.8 & 4.4  & 0.7 & 4.9 & 9.3 \\
                    & Ours       &     \textbf{23.8}	&\textbf{42.7}&	\textbf{23.8}&	\textbf{9.9}&	\textbf{24.9}&	\textbf{34.6} & \textbf{21.4}&	\textbf{39.2}&	\textbf{21.2}&	\textbf{6.2}&	\textbf{20.5}&	\textbf{33.2}  \\      
                    \bottomrule
\end{tabular}
}
\vspace{-0.13in}
\caption{\textbf{Complete Table for Few-shot Instance Segmentation on COCO}. All models use ImageNet \cite{imagenet_cvpr09} pretrained ResNet-50 \cite{he2016deep} as the backbone.} 
\label{tab:coco-few-segm-full}
\vspace{-0.1in}
\end{table*}
\vspace{-0.1in}
\section{Comparison to Few-shot Instance Segmentation}
\vspace{-0.05in}
\label{sec:cocoseg}
As described in Section \ref{exp:vocexp} of the main paper, we analyse the performance of our proposed approach on the task of Few-shot Instance Segmentation. Please refer to Section \ref{exp:vocexp} for task and dataset specifications. Similar to \cite{Yan_2019_ICCV}, we choose ResNet-50 \cite{he2016deep} as the backbone and use an additional segmentation head (as described in Section \ref{sec:segmentation} of the supplementary). The $k$-shot tasks are sampled following \cite{Yan_2019_ICCV} with $k \in \{0,5,10,20\}$, and we follow the standard evaluation metrics on COCO \cite{he2017mask}. The complete results are shown in Table \ref{tab:coco-few-segm-full}. Our approach consistently improves over \cite{Yan_2019_ICCV}, demonstrating that our approach is not limited to bounding boxes and is able to generalize over the type of downstream structured label, including segmentation mask.

% As can be seen from Table~\ref{tab:coco-few-segm-full} we get a significant boost in performance with $k=5$. However, the improvement obtained by going to $k=20$ is comparatively less significant. This is consistent with observations made in \cite{Yan_2019_ICCV}. 

\vspace{-0.08in}
\section{Weakly-Supervised Object Detection}
\label{exp:weaklyexp}
\vspace{-0.08in}
\noindent\textbf{Dataset.} 
For completeness, we compare our approach against existing weakly-supervised object detection approaches. We evaluate our approach on VOC 2007 \cite{everingham2010pascal} which consists of a trainval set of $5011$ images for training and $4951$ images for test, keeping in line with the prior related works \cite{bilen2016weakly, tang2017multiple, tang2018pcl, wang2018collaborative, arun2019dissimilarity}. As we assume instance-level supervision for \emph{base} classes in the dataset, we report performance on the \emph{novel} classes for three different splits specified in Section \ref{exp:vocexp} and \cite{wang2020frustratingly, Yan_2019_ICCV}. Note that, similar to the baselines, we assume \emph{zero} instance-level supervision for \emph{novel} classes (\ie $k=0$). 

% We compare our results with the state-of-the-art on the novel classes. 
% We use OICR \cite{tang2017multiple} to model weak-supervision and use base detection data to train base detectors. We use ImageNet pre-trained VGG16 \cite{simonyan2014very} as the backbone to remain consistent with the prior works (cite).

% Discuss variants of our model and relevant hyperparams here.

% \textbf{Baselines.} We report results of related baselines \cite{bilen2016weakly, tang2017multiple}.
% Comparison to OICR \cite{tang2017multiple} is most meaningful as this is the weakly-supervised branch in our model.
% % and in particular, we choose OICR \cite{tang2017multiple} as the weakly-supervised branch in our model. 
% We report figures from both the published \cite{tang2017multiple} and our re-implemented version of OICR, and we also include latest state-of-the-art method \cite{arun2019dissimilarity} in weakly-supervised object detection. For an extensive list, please see Table 1 in \cite{arun2019dissimilarity}.

% Please add the following required packages to your document preamble:
% \usepackage{booktabs}
% \begin{wraptable}{r}{0.45\textwidth}
\begin{table*}
% \vspace{-5mm}
\centering
\setlength\tabcolsep{2.5pt}
\resizebox{\textwidth}{!}{
\begin{tabular}{@{}ccccccc|cccccc|cccccc@{}}
\toprule
& \multicolumn{6}{c|}{Novel split 1}       & \multicolumn{6}{c|}{Novel split 2}         & \multicolumn{6}{c}{Novel split 3}  \\
Method      &bird&	bus&	cow&	mbike&	sofa&	mean& aero&	bottle&	cow&	horse&	sofa&	mean&   boat	&cat&	mbike&	sheep&	sofa&	mean \\ \midrule
WSDDN \cite{bilen2016weakly}   &31.5&	64.5&	35.7&	55.6&	40.7&	45.6& 39.4&	12.6&	35.7&	34.4&	40.7&	32.6& 16.3 &	42.6 & 55.6 & 30.2&	40.7&	37.1\\
OICR \cite{tang2017multiple} &31.1	&65.1	&44.7	&65.5	&46.9	&50.7& 58.0	&13.0	&44.7	&37.8	&46.9	&40.1   & 19.4&	28.4&	65.5&	41.7&	46.9&	40.4 \\
PCL \cite{tang2018pcl} &39.3	&62.9	&52.5	&67.7	&57.5	&56.0 &54.4	&15.7	&52.5	&39.3	&57.5	&43.9 &19.2 &	30.0&	67.7&	46.6&	57.5&	44.2\\
PredNet \cite{arun2019dissimilarity} &52.8 &\textbf{74.5}&	53.0&	70.8&	\textbf{60.7}&	62.4& 66.7	&24.7 &53.0 &\textbf{69.9} &	\textbf{60.7}&	55.0& \textbf{31.4}	& 67.3 &	70.8&	54.6&	\textbf{60.7}&	57.0\\ 
Wetectron \cite{ren2020instance} &\textbf{57.0} &69.1 &73.2 &\textbf{77.7} &53.8 &\textbf{66.2} &\textbf{68.8}	&\textbf{28.9}&	73.2&	54.4&	53.8&	\textbf{55.8}& 27.7	& 67.0 &	\textbf{77.7}&	\textbf{64.1}&	53.8&	58.1\\ \midrule
UniT + OICR (Ours) &45.5&	71.8&	\textbf{75.1}&	74.0&	52.7&	63.8 &  64.0&	17.6&	\textbf{73.8}&	59.9&	54.4&	53.9       &30.8 &	\textbf{71.7}&	74.9	&63.4	&55.1&	\textbf{59.2}  \\ \bottomrule
\end{tabular}
}
\vspace{-3mm}
\caption{\textbf{Comparison to Weakly-supervised methods.} Comparison on the three \emph{novel} splits described in Section \ref{exp:vocexp} and \cite{Yan_2019_ICCV, wang2020frustratingly}. All methods use VGG-16 \cite{simonyan2014very} as the backbone.}
\label{tab:weak-sup-voc07}

% \end{wraptable}
\end{table*}
\begin{figure*}[t]
\vspace{-0.1in}
\centering
\includegraphics[width=0.8\textwidth]{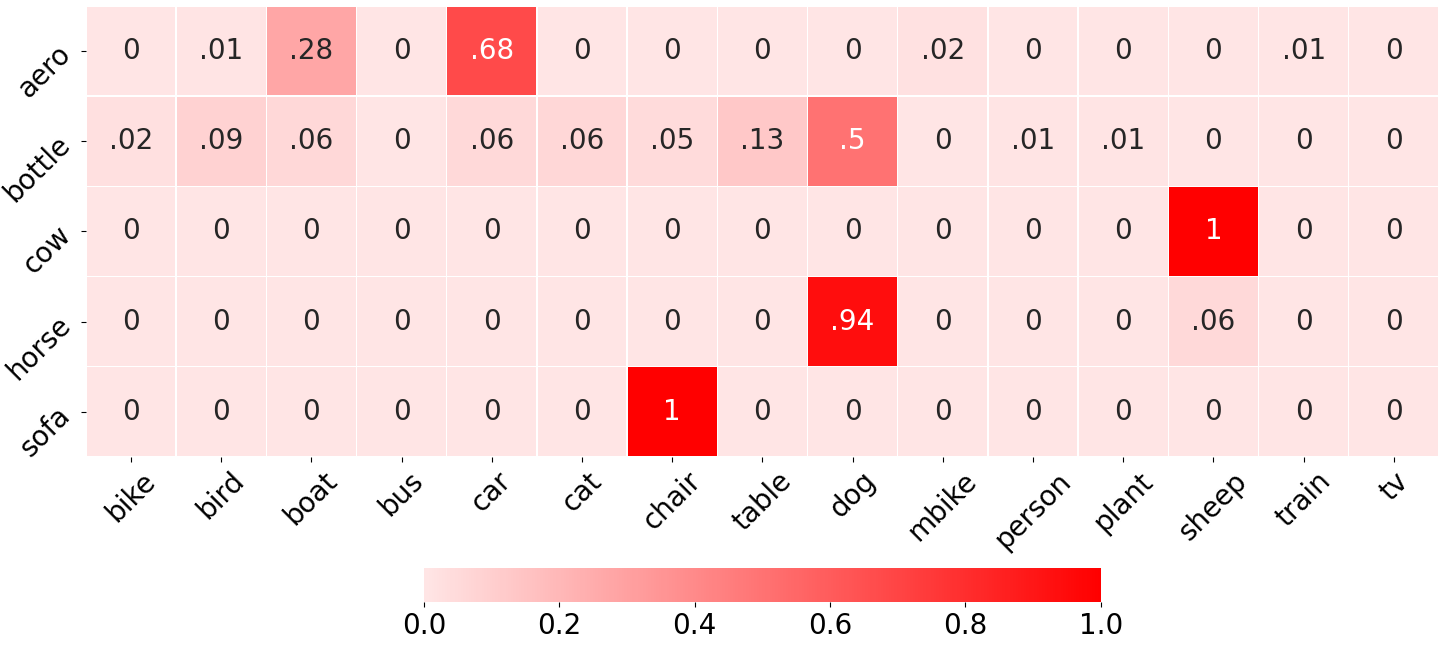}
\caption{\textbf{\emph{Normalized} lingual similarity matrix for the second \emph{novel} split in PASCAL VOC.} Note that $\mathbf{S}^{lin}$ is proposal-agnostic. Most of the similarities are intuitive and semantic -- {\tt sofa} is most similar to a {\tt chair}; {\tt horse} to a  {\tt dog} and a {\tt sheep}; {\tt cow} is similar to a {\tt sheep}; {\tt aeroplane} is related to other transportation vehicles like {\tt car} and {\tt boat}. A notable departure is a {\tt bottle} which has no closely related categories among {\em base} classes, resulting in less interpretable similarity and transfer.}
\label{fig:lingualsim}
\end{figure*}
\begin{table*}[t]
\centering
\setlength\tabcolsep{2.5pt}
% \resizebox{\textwidth}{!}{%
\begin{tabular}{@{}ccccccc|cccccc@{}}
\toprule
Method                  & \multicolumn{6}{c|}{$k=0$}       & \multicolumn{6}{c}{$k=5$}          \\ \midrule
                        & bird & bus  & cow  & mbike & sofa & mean & bird & bus  & cow  & mbike & sofa & mean  \\ \midrule
weak                      & 58.1 & 73.5 & 70.4 & 68.5  & 49.1 & 63.9 & 59.6 & 75.9 & 72.7 & 71.8 & 55.1 & 67.0  \\
weak $+\; \text{avg} (\Delta)$           & 57.8 & 73.5 & 71.2 & 68.3  & 47.8 & 63.7 & 60.0 & 76.0 & 73.6 & 71.2 & 54.2 & 67.0 \\
weak $+\; \mathbf{S}^{lin}_{cls}$     & 55.0 & 74.2 & 73.6 & 70.8  & 46.6 & 64.1 & 58.3 & 76.8 & 76.1 & 73.5 & 52.1 & 67.4\\
weak $+\; \mathbf{S}^{lin}_{cls} + \mathbf{S}^{vis}_{cls}$  & 56.2 & 74.6 & 73.9 & 71.2  & 48.8 & 64.9 & 59.3 & 77.1 & 77.2 & 73.9 & 52.0 & 67.9\\
weak $+\; \mathbf{S}^{lin}_{cls,reg} + \mathbf{S}^{vis}_{cls,reg}$                 & \textbf{69.9} & \textbf{83.4} & \textbf{86.1} & \textbf{81.1} & \textbf{57.8} & \textbf{75.7} & \textbf{69.8} & \textbf{84.0} & \textbf{86.2} &\textbf{81.3} & \textbf{59.0} & \textbf{76.1} \\ \bottomrule
% \vspace{-0.3in}
\end{tabular}%
% }
\caption{\textbf{Ablation study on the VOC 07 + 12 dataset. }Please refer to Section \ref{sec:ablation} for model definitions. We report AP$_{50}$ on the \emph{first} split in \cite{Kang_2019_ICCV}, and show results on zero-shot $(k=0)$ and few-shot $(k=5)$.}
\label{tab:ablationtable-voc}
\end{table*}
\begin{table*}[t]
% \vspace{-0.06in}
\centering
\setlength\tabcolsep{2.5pt}
\renewcommand{\arraystretch}{1.2}
% \resizebox{\textwidth}{!}{%
\scalebox{0.9}{
\begin{tabular}{@{}c@{\hskip 0.3in}cccccc@{\hskip 0.3in}cccccc@{}}
\toprule
                            & \multicolumn{6}{c}{Box}             & \multicolumn{6}{c}{Mask}            \\ \midrule
Method     & AP  & AP$_{50}$ & AP$_{75}$ & AP$_{S}$ & AP$_M$ & AP$_L$ & AP  & AP$_{50}$ & AP$_{75}$ & AP$_S$ & AP$_M$ & AP$_L$ \\ \midrule
weak & 10.7 &	28.3 &	5.1 &	5.3 &	12.9 &	14.3 & 4.7 &	17.6 &	1.1 & 2.1 &	6.0 &	6.0\\
weak $+\; \text{avg} (\Delta)$ &10.7& 	28.3&	5.1&	5.3 &	12.9 &	14.2 & 4.7	&17.5 &	1.1 &	2.0 &	6.0 &	6.0 \\
weak $+\; \mathbf{S}^{lin}_{cls}$ &11.4 &	29.7 &	5.7 &	6.3 &	13.2 &	14.6 & 5.0 &	18.5 &	1.0 &	2.3 &	6.0 &	6.1 \\
weak $+\; \mathbf{S}^{lin}_{cls} + \mathbf{S}^{vis}_{cls}$ &11.7	&29.9 &	6.0 &	6.4 &	13.3 &	14.7 & 5.2 &	18.7 &	1.2 &	2.4 &	6.1 &	6.1 \\
weak $+\; \mathbf{S}^{lin}_{cls,reg} + \mathbf{S}^{vis}_{cls,reg}$ & 20.2 &	36.8 &	19.5 & 	8.5 &	20.9 & 	28.9 & 8.5 &	25.0 &	4.5 &	2.8 &	8.7 &	12.0 \\
weak $+\; \mathbf{S}^{lin}_{cls,reg,seg} + \mathbf{S}^{vis}_{cls,reg,seg}$ & \textbf{20.2} &	\textbf{36.8} &	\textbf{19.5} &	\textbf{8.5} &\textbf{20.9} &	\textbf{28.9} & \textbf{17.6} & \textbf{32.7} &	\textbf{17.0} &	\textbf{5.6} &	\textbf{17.7} &	\textbf{27.7}\\ 
                    \bottomrule
\end{tabular}
}
\vspace{-0.1in}
\caption{\textbf{Ablation study on the MSCOCO dataset. }Please refer to Section \ref{sec:ablation} for model definitions. The results show the \emph{zero} shot ($k=0$) performance of the models.}
\label{tab:ablationtable}
\vspace{-0.1in}
\end{table*}

\begin{table*}[h]
	\begin{adjustbox}{center}
	\setlength{\tabcolsep}{1.2pt}
	\fontsize{7.6}{10.5}{\selectfont
		\begin{tabular}{c |c| c c c c c c | c c c c c c c c c c c c c c c c|c}
			\hline
			\multicolumn{2}{c|}{}&\multicolumn{6}{c|}{Novel classes}&\multicolumn{16}{c|}{Base classes}&\multicolumn{1}{c}{\multirow{2}*{mAP}}\cr\cline{1-24}
			Shot & Method & bird & bus & cow & mbike & sofa & mean & aero   & bike & boat & bottle & car & cat & chair & table & dog & horse & person & plant & sheep & train & tv 
			& mean \cr\cline{1-23} \hline
			\multirow{2}*{0}
			& Wang \etal~\cite{wang2020frustratingly} &- &- &- &- &- &9.0 & - &- &- &- &- &- &- &- &- &- &- &- &- &- &- &\textbf{80.8} &62.9\\
			& UniT (Ours) \etal~\cite{wang2020frustratingly} &69.9 &	83.4 &	86.1 &	81.1 &	57.8 &	\textbf{75.6} & 82.0 &84.7&   70.0&    69.2&    87.9&    88.4&    60.5&    71.3&    84.8&    86.1&    84.5&    54.0&    82.2&    85.1&    78.6 & 78.0&\textbf{77.3}\\
			\hline
			\multirow{5}*{3}
			&Joint: FRCN \cite{Yan_2019_ICCV} &13.7 &0.4 &6.4 &0.8 &0.2 &4.3 &75.9 &80.0 &65.9 &61.3 &85.5 &86.1 &54.1 &68.4 &83.3 &79.1 &78.8 &43.7 &72.8 &80.8 &74.7 &72.7 &55.6 \\
			&Transfer: FRCN \cite{Yan_2019_ICCV} &29.1 &34.1 &55.9 &28.6 &16.1 &32.8 &67.4 &62.0 &54.3 &48.5 &74.0 &85.8 &42.2 &58.1 &72.0 &77.8 &75.8 &32.3 &61.0 &73.7 &68.6 &63.6 &55.9\\
			& Kang \etal~\cite{Kang_2019_ICCV} &26.1 &19.1 &40.7 &20.4 &27.1 &26.7 &73.6 &73.1 &56.7 &41.6 &76.1 &78.7 &42.6 &66.8 &72.0 &77.7 &68.5 &42.0 &57.1 &74.7 &70.7 &64.8 &55.2 \\
% 			&FRCN+ft &31.1 &24.9 &51.7 &23.5 &13.6 &29.0 &65.4 &56.4 &46.5 &41.5 &73.3 &84.0 &40.2 &55.9 &72.1 &75.6 &74.8 &32.7 &60.4 &71.2 &71.2 &61.4 &53.3\\
			&Yan \etal~\cite{Yan_2019_ICCV} &30.1 &44.6 &50.8 &38.8 &10.7 &35.0 &67.6 &70.5 &59.8 &50.0 &75.7 &81.4 &44.9 &57.7 &76.3 &74.9 &76.9 &34.7 &58.7 &74.7 &67.8 &64.8 &57.3 \\
			&Wang \etal~\cite{wang2020frustratingly}&- &- &- &- &- &44.7 & - &- &- &- &- &- &- &- &- &- &- &- &- &- &- &\textbf{79.1} &70.5\\ 
			&UniT (Ours) & \textbf{70.0} & \textbf{83.9}& \textbf{86.2}& \textbf{81.5}& \textbf{58.0} &\textbf{75.9}& \textbf{81.9}&  \textbf{84.7} &\textbf{69.0}& \textbf{68.9}& \textbf{87.9}& \textbf{88.1}& \textbf{60.4}& \textbf{71.3}& \textbf{84.7}& \textbf{86.2}& \textbf{84.4}& \textbf{54.2}& \textbf{82.0}& \textbf{84.7}& \textbf{78.8}& 77.8& \textbf{77.3}
			
			\\
			\hline
			\multirow{5}*{10}
			&Joint: FRCN \cite{Yan_2019_ICCV} &14.6 &20.3 &19.2 &24.3 &2.2 &16.1 
			&78.1 &80.0 &65.9 &64.1 &86.0 &87.1 &56.9 &69.7 &84.1 &80.0 &78.4 &44.8 &74.6 &82.7 &74.1 &73.8 &59.4\\
% 			&FRCN+ft &31.3 &36.5 &54.1 &26.5 &36.2 &36.9 &68.4 &75.2 &59.2 &54.8 &74.1 &80.8 &42.8 &56.0 &68.9 &77.8 &75.5 &34.7 &66.1 &71.2 &66.2 &64.8 &57.8 \\
			&Transfer: FRCN \cite{Yan_2019_ICCV} &40.1 &47.8 &45.5 &47.5 &47.0 &45.6 &65.7 &69.2 &52.6 &46.5 &74.6 &73.6 &40.7 &55.0 &69.3 &73.5 &73.2 &33.8 &56.5 &69.8 &65.1 &61.3 &57.4\\
			& Kang \etal~\cite{Kang_2019_ICCV} &30.0 &62.7 &43.2 &60.6 &39.6 &47.2 &65.3 &73.5 &54.7 &39.5 &75.7 &81.1 &35.3 &62.5 &72.8 &78.8 &68.6 &41.5 &59.2 &76.2 &69.2 &63.6 &59.5\\
			&Yan \etal~\cite{Yan_2019_ICCV} &52.5 &55.9 &52.7 &54.6 &41.6 &51.5 &68.1 &73.9 &59.8 &54.2 &80.1 &82.9 &48.8 &62.8 &80.1 &81.4 &77.2 &37.2 &65.7 &75.8 &70.6 &67.9 &63.8\\
			&Wang \etal~\cite{wang2020frustratingly} &- &- &- &- &- &56.0 & - &- &- &- &- &- &- &- &- &- &- &- &- &- &- &\textbf{78.4} &72.8\\
			&UniT (Ours) & \textbf{71.4} & \textbf{84.4} &    \textbf{86.3} &    \textbf{82.2} &    \textbf{59.2}&    \textbf{76.7}&    \textbf{82.0} &   \textbf{84.6} &   \textbf{68.9} &   \textbf{69.3} &   \textbf{87.7}&    \textbf{88.0}  &  \textbf{59.7}  &  \textbf{71.4} &\textbf{84.9} &\textbf{86.5} &  \textbf{84.6} &   \textbf{54.0} &   \textbf{81.2} &   \textbf{84.0} &   \textbf{78.3} & 77.7& \textbf{77.4 }      
			\\
			\hline
		\end{tabular}}
		\end{adjustbox}
	\caption{{\bf Base class performance on VOC.} AP and mAP on VOC2007 test set for novel classes and base classes of the first base/novel split. We evaluate the performance for 0/3/10-shot novel-class examples with FRCN under ResNet-101. 
	Note that as Wang \etal~\cite{wang2020frustratingly} do not report per-class performance numbers, we just show their reported aggregated performance. Additionally, all models use the \emph{same} amount of annotation for the \emph{base} classes.
		}		\label{tab:meta-rcnn-base}
	\end{table*}

\vspace{0.5em}
\noindent\textbf{Results.}
% \textcolor{blue}{Sofa is present in all splits. We do poorly on that which drags the mean down.}
Table \ref{tab:weak-sup-voc07} provides a summary of the results. Similar to \cite{bilen2016weakly, tang2017multiple, tang2018pcl, arun2019dissimilarity,ren2020instance}, we use a pre-trained proposal network (Selective Search \cite{uijlings2013selective}) and an ImageNet pretrained VGG-16 \cite{simonyan2014very} backbone for fair comparison. As we assume additional supervision for \emph{base} classes, this is not an equivalent comparison. However, for \emph{novel} classes, all methods have access to the \emph{same} data. Our results beat the strong baseline of \cite{arun2019dissimilarity} on $2$ out of $3$ novel splits, and provides comparable performance to the state-of-the-art \cite{ren2020instance}. Our significant improvement over OICR \cite{tang2017multiple}, which we build upon, on \emph{novel} classes (${\sim}35\%$ on average across three splits) highlights the effectiveness of our proposed transfer. We note that our approach is agnostic to the model architecture used for weak-supervision, and the model performance can be improved further if built on top of better weak detectors ({\em e.g.} \cite{arun2019dissimilarity, ren2020instance}). 

\section {Explanation of Annotation Budget Conversion Factor}
\label{sec:budget-details}
In Section \ref{exp:annotation} of the main paper, we perform a constraint annotation budget analysis to facilitate an equivalent comparison to existing few-shot detection works \cite{wang2020frustratingly, Kang_2019_ICCV}, while simultaneously analysing the relative importance of image-level annotations. To this end, for each value of $k$ instance-level annotations in the few-shot setup, we trained a variant of UniT that assumes only $7\times k$ image-level annotations for \emph{novel} classes, which is referred to as UniT$_{budget=k}$. Here we describe the rationale behind using this conversion factor of $7$ for the VOC dataset \cite{everingham2015pascal}, derived from annotation times reported in the literature.

\vspace{0.5em}
\noindent\textbf{Image-Level Annotation (20.0 sec/image).} As per \cite{Bearman_2019_ECCV} and \cite{papadopoulos2014training}, collecting image-level labels takes 1 second per class. As VOC \cite{everingham2015pascal} has $20$ object classes, generating an image-level annotation is expected to take $20$ seconds per image \cite{Bearman_2019_ECCV}.

\vspace{0.5em}
\noindent\textbf{Instance-Level Bounding Box Annotation (137.2 sec/image).} \cite{Bearman_2019_ECCV} reports an expected annotation time of $239.7$ seconds per image for VOC \cite{everingham2015pascal}. However, this estimate additionally assumes time required to obtain instance segmentations. As the methods in the few-shot detection domain \cite{wang2020frustratingly, Kang_2019_ICCV} that we compare against in Section \ref{exp:annotation} only use bounding box annotations to train their models, we modify this estimate of $239.7$ seconds to get a more accurate conversion factor.

There are 1.5 object classes per image on average in the VOC 2012 dataset \cite{everingham2015pascal}. As mentioned in \cite{Bearman_2019_ECCV}, it takes 1 second to annotate every object class that is not present (\ie obtain an image-level ``no object'' label). This amounts to $20 - 1.5 = 18.5$ seconds of labeling time. In addition, there are 2.8 object instances on average in the VOC dataset \cite{everingham2015pascal}. As mentioned in \cite{su2012crowdsourcing}, if we use their efficient system to crowdsource annotations, the median time to get a bounding box annotation is $42.4$ seconds (note that the average time is much higher at $88.0$ seconds). Therefore, it is expected to take $2.8 \times 42.4 = 118.7$ seconds to obtain bounding box annotations per image in VOC. Thus, the total annotation time is: $18.5 + 118.7 = 137.2$ seconds per image.

\vspace{0.5em}
\noindent It can be seen that obtaining instance-level annotations is approximately $137.2/20 \approx 7$ times more time consuming. Hence, we use a conversion factor of $7$ for our experiments in Section \ref{exp:annotation}. Also, according to \cite{papadopoulos2017training}, this estimate of $42.4$ seconds per bounding box is \emph{conservative}. It doesn't take into account the errors encountered during crowdsourcing (\eg some boxes being rejected and needing to be redrawn). Therefore, we expect this conversion factor of $7$ to be higher in practice.

\section{Ablation study}
\label{sec:ablation}
\vspace{-0.05in}
Please refer to Section \ref{exp:vocexp} of the main paper for a detailed explanation of task setup. We perform ablation over the terms used in Equations (\ref{eq:cls}), (\ref{eq:reg}), and (\ref{eq:seg}) of the main paper on the \emph{novel} classes for the VOC \cite{lin2014microsoft} and MSCOCO \cite{lin2014microsoft} datasets. We show the benefit of using our proposed transfer both in the zero-shot ($k=0$) and the few-shot ($k>0$) setting. We first describe the ablated variants of our model, and then discuss the results. The results are summarized in Table \ref{tab:ablationtable-voc} for VOC and Table \ref{tab:ablationtable} for MSCOCO.

% The results are summarized in Table \ref{tab:ablationtable}. Note that we do not use any instance-level annotations for the \emph{novel} classes, {\em i.e.}, we report results for the semi-supervised zero-shot ($k=0$) setting.
%

%
We start by forming detectors using only the weakly-supervised branch $f_{\mathbf{W}^{weak}}$ (denoted as ``weak'' in Tables \ref{tab:ablationtable-voc} and \ref{tab:ablationtable}), and progressively add refinement terms to observe their impact on detection/segmentation performance. We then incorporate the transfer from the \emph{base} classes $f_{\Delta \mathbf{\W}_{base}^{cls}}$ into the weak detector (see Equation \ref{eq:cls} in the main paper). For each \emph{novel} class, we first compare to a simple baseline approach: averaging over all the \emph{base} classes (denoted by weak $+\; \text{avg} (\Delta)$). We note that a na\"ive averaging doesn't provide any performance improvements, suggesting the need for a more informed transfer between {\em base} and {\em novel} classes.
% and averaging over top-5 most similar classifiers (denoted by weak $+\;\text{top-5}(\Delta)$). For each \emph{novel} class, similar to LSDA \cite{hoffman2014lsda}, the top-5 most similar classifiers are computed using the inner-product between the weights of $f_{\Delta\mathbf{W}_{novel}^{cls}}$ and $f_{\Delta\mathbf{W}_{base}^{cls}}$. 
% We note that top-5 (row 3) performs better than na\"ive averaging (row 2), which shows that an informed similarity measure between {\em base} and {\em novel} classes leads to better performance.
We then explore the role of proposed similarity matrices, detailed in Section \ref{sec:similarity} of the main paper. 
% Furthermore, we start weighing the transfer from the \emph{base} classes $f_{\Delta \mathbf{\W}_{base}^{cls}}$ using our proposed similarity matrices detailed in Section \ref{sec:similarity} of the main paper. 
The similarity matrix between \emph{base} and \emph{novel} classes can be decomposed into two components: lingual similarity $\mathbf{S}^{lin}$ and visual similarity $\mathbf{S}^{vis}(\mathbf{z})$. We analyse the impact of using the aforementioned similarities in obtaining category-aware classifiers, regressors, and segmentors. Following the terms in Eq. (\ref{eq:cls}), (\ref{eq:reg}), and (\ref{eq:seg}) of the main paper, we define ablated variants of our final model.
\begin{itemize}
    \item ``weak $+\;\mathbf{S}^{lin}_{cls}$'' is the model where-in the category-aware classifier for the \emph{novel} classes is obtained by using only the lingual similarity, and the category-aware regressor is fixed to predict zeros (\emph{i.e.} the model uses the output of the category-agnostic Fast-RCNN regressor $\mathbf{rbox}$). As there is no estimate for the \emph{novel} mask head, we predict a uniform mask over the selected bounding box region.
    
    \item ``weak $+\;\mathbf{S}^{lin}_{cls}+\mathbf{S}^{vis}_{cls}$'' is defined as the model where-in the category-aware classifier for the \emph{novel} classes is obtained by using both lingual and visual similarities (Eq. (\ref{eq:cls})), and the category-aware regressor is fixed to predict zeros. As there is no estimate for the \emph{novel} mask head, we predict a uniform mask over the selected bounding box region.
    
    \item  ``weak $+\;\mathbf{S}^{lin}_{cls,reg}+\mathbf{S}^{vis}_{cls,reg}$'' is defined as the model where-in both the category-aware classifier and the category-aware regressor for the \emph{novel} classes is obtained by using lingual and visual similarities (Eq. (\ref{eq:cls})) and (\ref{eq:reg})). As there is no estimate for the \emph{novel} mask head, we predict a uniform mask over the selected bounding box region. For experiments in Table \ref{tab:ablationtable-voc}, this is the complete UniT model.
    
    \item  Finally, ``weak $+\;\mathbf{S}^{lin}_{cls,reg,seg}+\mathbf{S}^{vis}_{cls,reg,seg}$'' is defined as the model where-in the category-aware classifier, the category-aware regressor, and the category-aware segmentor for the \emph{novel} classes is obtained by using lingual and visual similarities (Eq. (\ref{eq:cls}), (\ref{eq:reg}), and (\ref{eq:seg})). For experiments in Table \ref{tab:ablationtable}, this is the complete UniT model.   
\end{itemize}
% Finally, in order to understand the impact of lingual similarity on both the category-aware classifiers and regressors, we define ``weak $+\;\mathbf{S}^{lin}$'' as the model that uses only lingual similarity in Eq.(3) and (4) of the main paper. The ``Final Model'' in Table \ref{tab:ablationtable} uses both similarities to obtain category-aware regressors and classifiers for the \emph{novel} classes. 
% The ablation clearly highlights importance of all terms in our model. 
From Table \ref{tab:ablationtable-voc} it can be seen that adding each of our terms improves model performance. Particularly, transferring information from \emph{base} regressors to \emph{novel} regressors (``weak $+\;\mathbf{S}^{lin}_{cls,reg}+\mathbf{S}^{vis}_{cls,reg}$'') provides a significant boost. We additionally show that this trend holds even when the models are fine-tuned on few-shot ($k=5$) examples for \emph{novel} classes. Table \ref{tab:ablationtable} further highlights the effectiveness of our proposed transfer on segmentation masks (``weak $+\;\mathbf{S}^{lin}_{cls,reg,seg}+\mathbf{S}^{vis}_{cls,reg,seg}$''). Note that the final two lines in Table \ref{tab:ablationtable} only differ in mask AP performance as the regression transfer terms are identical in both the ablated models.

\begin{figure*}[t]
% \captionsetup[subfigure]{labelformat=empty}
\begin{subfigure}{.25\textwidth}
\centering
\includegraphics[width=0.9\textwidth, height=3.5cm]{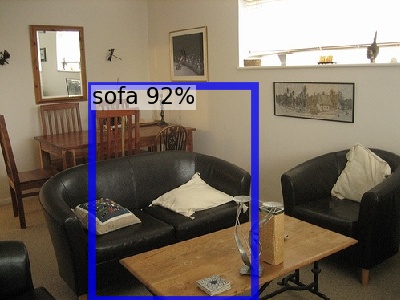}
\end{subfigure}%
\begin{subfigure}{.25\textwidth}
\centering
\includegraphics[width=0.9\textwidth, height=3.5cm]{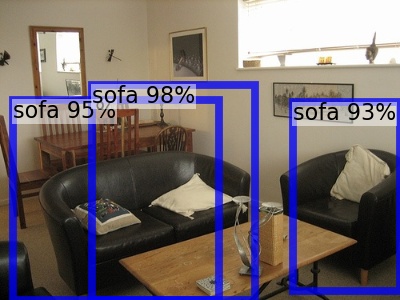}
\end{subfigure}%
\begin{subfigure}{.25\textwidth}
\centering
\includegraphics[width=0.9\textwidth, height=3.5cm]{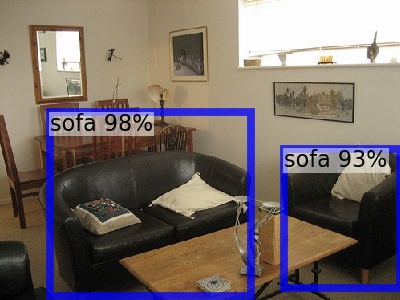}
\end{subfigure}%
\begin{subfigure}{.25\textwidth}
\centering
\includegraphics[width=0.9\textwidth, height=3.5cm]{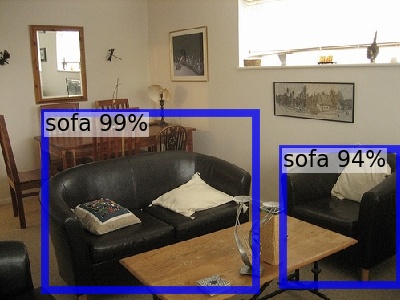}
\end{subfigure}
\begin{subfigure}{.25\textwidth}
\centering
\includegraphics[width=0.9\textwidth, height=4.5cm]{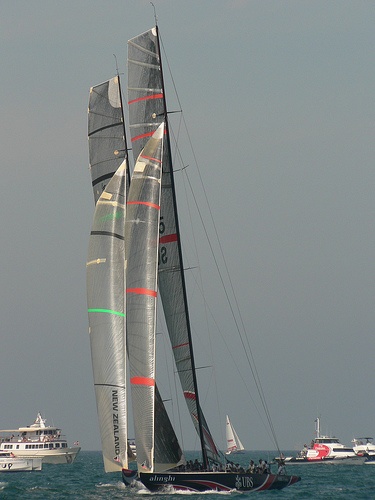}
\end{subfigure}%
\begin{subfigure}{.25\textwidth}
\centering
\includegraphics[width=0.9\textwidth, height=4.5cm]{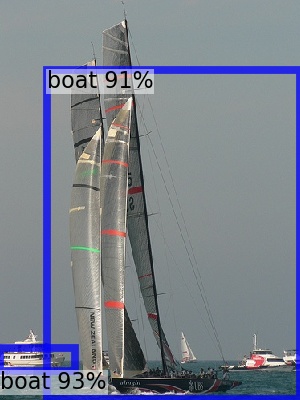}
\end{subfigure}%
\begin{subfigure}{.25\textwidth}
\centering
\includegraphics[width=0.9\textwidth, height=4.5cm]{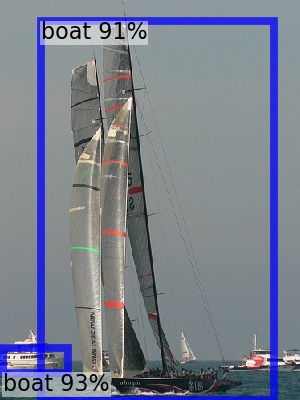}
\end{subfigure}%
\begin{subfigure}{.25\textwidth}
\centering
\includegraphics[width=0.9\textwidth, height=4.5cm]{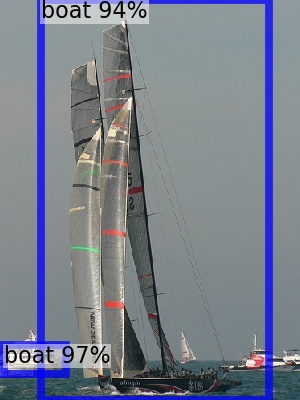}
\end{subfigure}
\begin{subfigure}{.25\textwidth}
\centering
\includegraphics[width=0.9\textwidth, height=3.5cm]{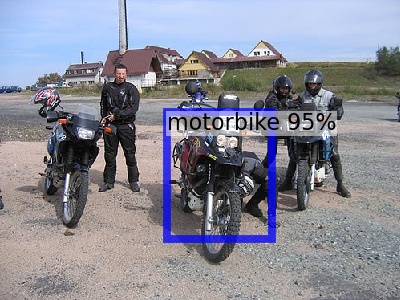}
\end{subfigure}%
\begin{subfigure}{.25\textwidth}
\centering
\includegraphics[width=0.9\textwidth, height=3.5cm]{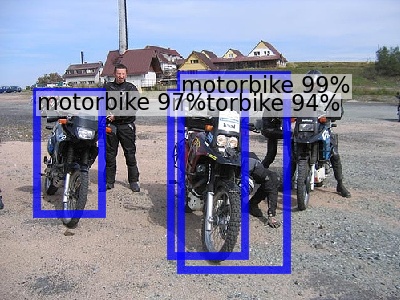}
\end{subfigure}%
\begin{subfigure}{.25\textwidth}
\centering
\includegraphics[width=0.9\textwidth, height=3.5cm]{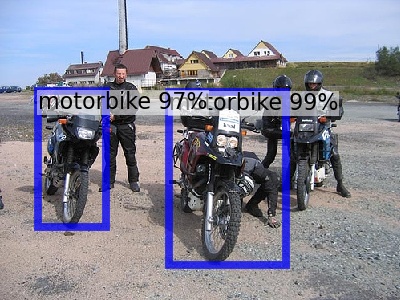}
\end{subfigure}%
\begin{subfigure}{.25\textwidth}
\centering
\includegraphics[width=0.9\textwidth, height=3.5cm]{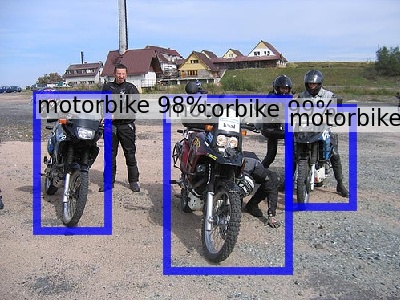}
\end{subfigure}
\begin{subfigure}{.25\textwidth}
\centering
\includegraphics[width=0.9\textwidth, height=3.5cm]{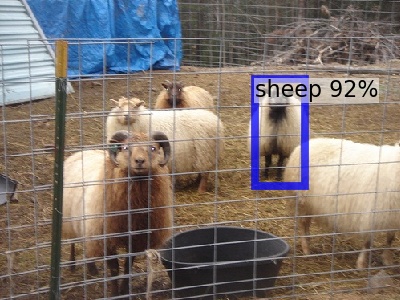}
\end{subfigure}%
\begin{subfigure}{.25\textwidth}
\centering
\includegraphics[width=0.9\textwidth, height=3.5cm]{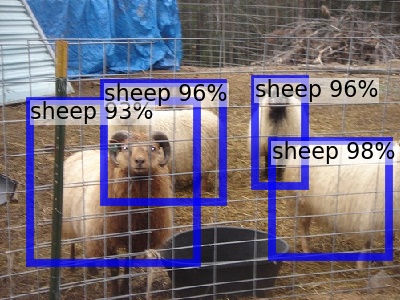}
\end{subfigure}%
\begin{subfigure}{.25\textwidth}
\centering
\includegraphics[width=0.9\textwidth, height=3.5cm]{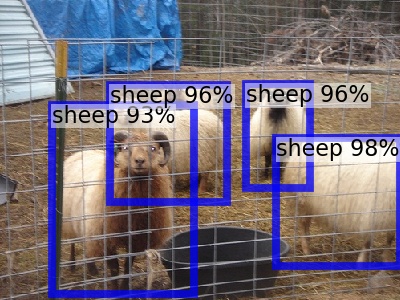}
\end{subfigure}%
\begin{subfigure}{.25\textwidth}
\centering
\includegraphics[width=0.9\textwidth, height=3.5cm]{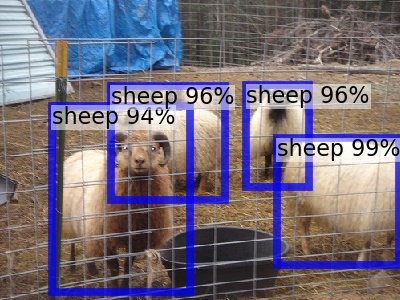}
\end{subfigure}
\begin{subfigure}{.25\textwidth}
\centering
\includegraphics[width=0.9\textwidth, height=3.5cm]{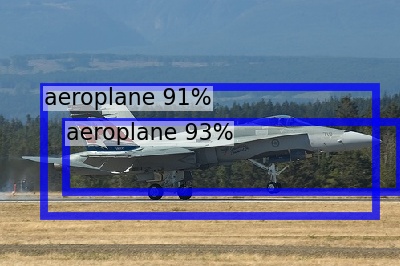}
\caption{}
\end{subfigure}%
\begin{subfigure}{.25\textwidth}
\centering
\includegraphics[width=0.9\textwidth, height=3.5cm]{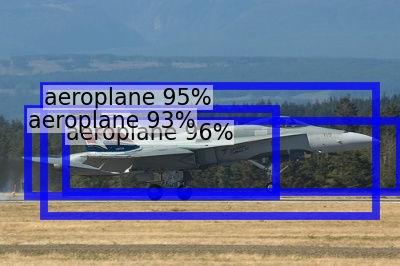}
\caption{}
\end{subfigure}%
\begin{subfigure}{.25\textwidth}
\centering
\includegraphics[width=0.9\textwidth, height=3.5cm]{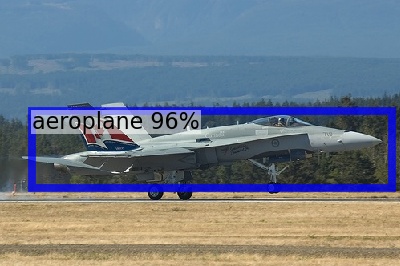}
\caption{}
\end{subfigure}%
\begin{subfigure}{.25\textwidth}
\centering
\includegraphics[width=0.9\textwidth, height=3.5cm]{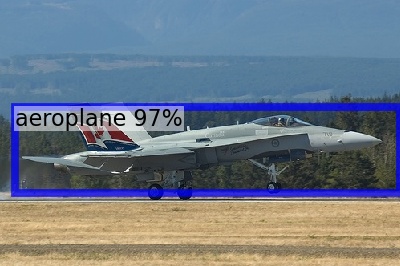}
\caption{}
\end{subfigure}
\caption{\textbf{Qualitative Visualizations for the Ablation Study.} (a) refers to the ``weak'' model, (b) refers to ``weak $+\; \mathbf{S}^{lin}_{cls} + \mathbf{S}^{vis}_{cls}$'', (c) refers to the ``weak $+\; \mathbf{S}^{lin}_{cls,reg} + \mathbf{S}^{vis}_{cls,reg}$'' with $k=0$, and (d) refers to the ``weak $+\; \mathbf{S}^{lin}_{cls,reg} + \mathbf{S}^{vis}_{cls,reg}$'' after being trained on $k=5$ shots. Section \ref{sec:ablation} provides a detailed description of these models.}
\label{suppfig:vocviz}
\end{figure*}

Figure \ref{suppfig:vocviz} provides qualitative examples to further highlight the impact of using our proposed transfer from \emph{base} to \emph{novel} classes. Column (a) in Figure \ref{suppfig:vocviz} refers to ``weak'', column (b) refers to ``weak $+\; \mathbf{S}^{lin}_{cls} + \mathbf{S}^{vis}_{cls}$'', column (c) refers to the ``weak $+\; \mathbf{S}^{lin}_{cls,reg} + \mathbf{S}^{vis}_{cls,reg}$'' with $k=0$, and column (d) refers to the ``weak $+\; \mathbf{S}^{lin}_{cls,reg} + \mathbf{S}^{vis}_{cls,reg}$'' after being trained with $k=5$ shots. It can be seen that the ``weak'' model either fails to identify all objects or doesn't generate high-probability proposals for the desired objects (column (a)). ``weak $+\; \mathbf{S}^{lin}_{cls} + \mathbf{S}^{vis}_{cls}$'' improves performance by generating a bunch of reasonable proposals (column (b)). ``weak $+\; \mathbf{S}^{lin}_{cls,reg} + \mathbf{S}^{vis}_{cls,reg}$'' further refines the proposals to obtain accurate bounding boxes for the objects (column (c)). Finally, fine-tuning on $k=5$ shots improves the bounding box confidence and slightly refines the predictions (column (d)).

\vspace{-0.05in}
\section{Analysis of Similarity Matrices}
\vspace{-0.05in}
\label{sec:similaritymatrix}

As described in Section \ref{sec:approach} of the main paper, the key contribution of our approach is the ability to semantically decompose the classifiers, detectors and segmentors of \emph{novel} classes into their base classes’ counterparts. To this end, we define a proposal-aware similarity $\mathbf{S}(\mathbf{z}) \in \mathbb{R}^{|C_{novel}|\times |C_{base}|}$, which is further  decomposed into two components: lingual $\mathbf{S}^{lin}$ and visual $\mathbf{S}^{vis}(\mathbf{z})$ similarity. Please refer to Section \ref{sec:similarity} of the main paper on details pertaining to how these similarities are computed.

\begin{figure*}[t]
\centering
\begin{subfigure}{\textwidth}
    \begin{subfigure}{.3\textwidth}
    \centering
    \includegraphics[width=\textwidth]{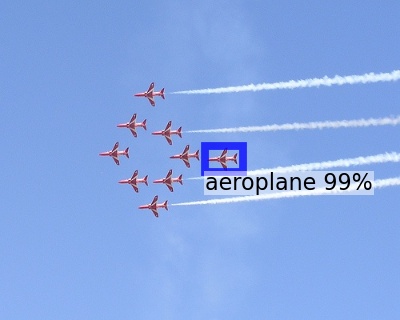}
    \end{subfigure}%
    \begin{subfigure}{.7\textwidth}
    \centering
    \includegraphics[width=0.97\textwidth]{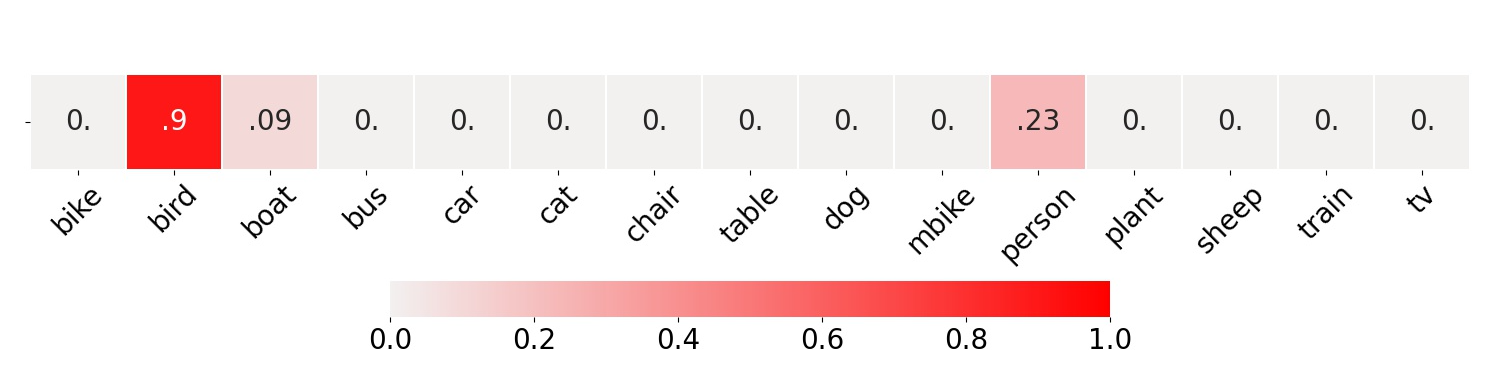}
    \end{subfigure}
    \caption{Complementary to $\mathbf{S}^{lin}$ that assigns weights to classes \texttt{boat} and \texttt{car}, $\mathbf{S}^{vis}(\mathbf{z})$ is additionally able to capture that an \texttt{aeroplane} flying in the sky shares some visual characteristics with the class \texttt{bird}.}
\end{subfigure}\par\bigskip
\begin{subfigure}{\textwidth}
    \begin{subfigure}{.3\textwidth}
    \centering
    \includegraphics[width=\textwidth]{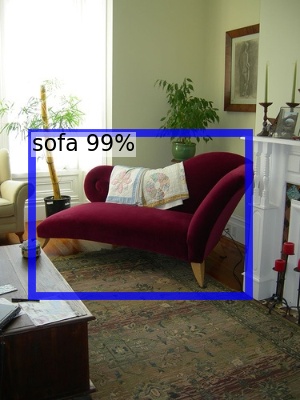}
    \end{subfigure}%
    \begin{subfigure}{.7\textwidth}
    \centering
    \includegraphics[width=0.97\textwidth]{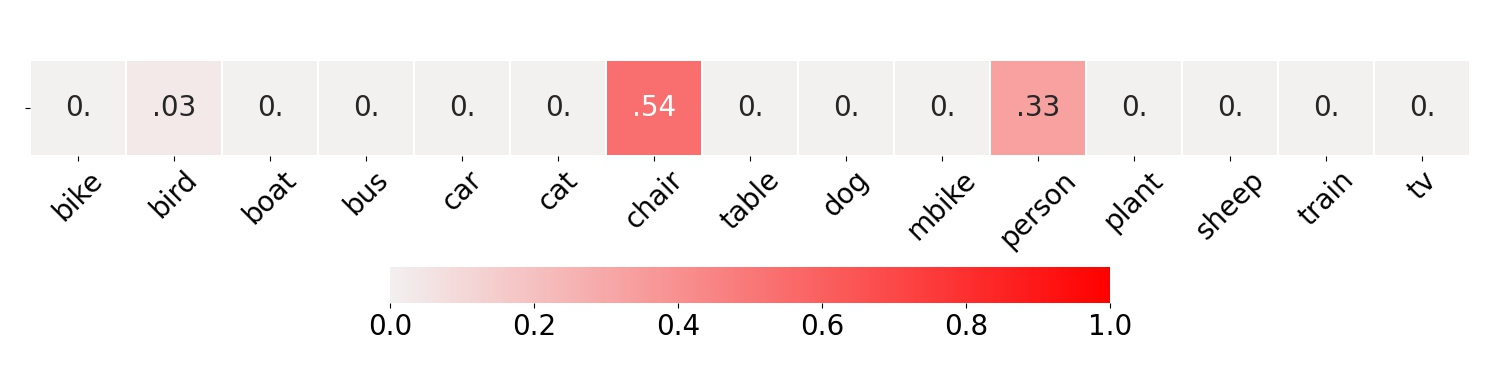}
    \end{subfigure}
    \caption{Complementary to $\mathbf{S}^{lin}$ that gives a large weight to the class \texttt{chair}, $\mathbf{S}^{vis}(\mathbf{z})$ is additionally able to capture that there is a high correlation between the class \texttt{person} and the class \texttt{sofa}. This follows the common observation that people are likely to be sitting on a sofa.}
\end{subfigure}\par\bigskip
\begin{subfigure}{\textwidth}
    \begin{subfigure}{.3\textwidth}
    \centering
    \includegraphics[width=\textwidth]{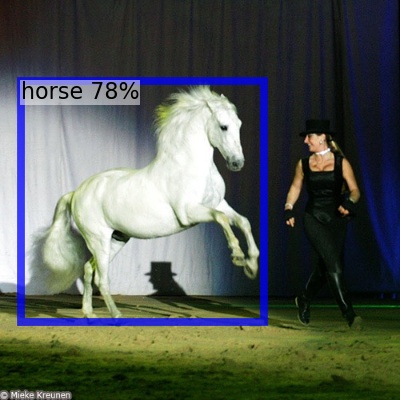}
    \end{subfigure}%
    \begin{subfigure}{.7\textwidth}
    \centering
    \includegraphics[width=0.97\textwidth]{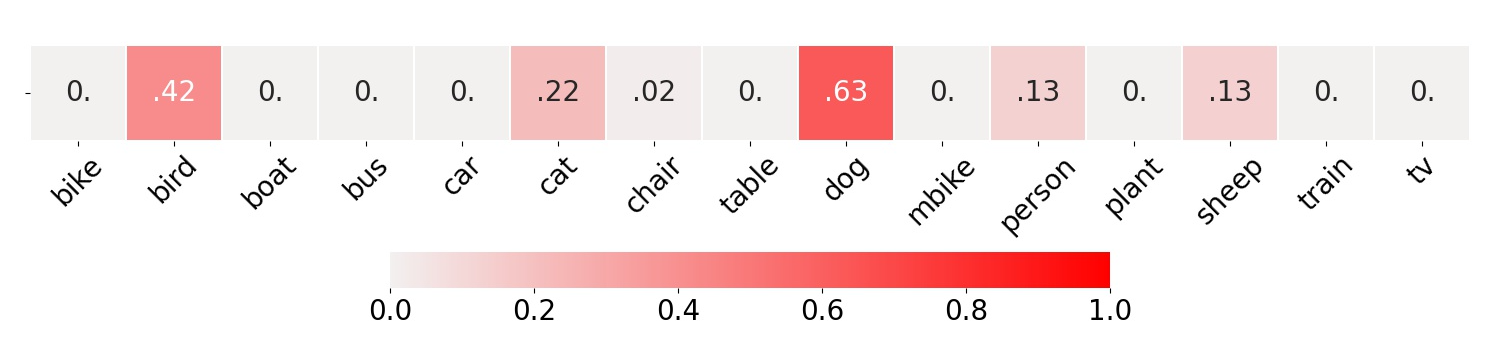}
    \end{subfigure}
    \caption{Complementary to $\mathbf{S}^{lin}$ that gives a large weight to the class \texttt{dog}, $\mathbf{S}^{vis}(\mathbf{z})$ is additionally able to capture that the class \texttt{horse} is visually similar to other animal classes \texttt{bird}, \texttt{cat}, and \texttt{sheep}. Additionally, it captures a correlation with the class \texttt{person}, which follows from the observation that humans usually ride horses.}
\end{subfigure}
\caption{\textbf{ Visual similarity for the second \emph{novel} split in PASCAL VOC.} The input proposal $\mathbf{z}$ is highlighted in \textcolor{blue}{{\bf blue}}. The visual similarity captures complementary information to the lingual similarity.}
\vspace{-0.2in}
\label{fig:visualsim}
\end{figure*}

We qualitatively visualize these similarity matrices to highlight the intuitive semantics learned by our proposed model. Figure \ref{fig:lingualsim} shows the $\emph{normalized}$ lingual similarity matrix $\mathbf{S}^{lin} \in \mathbb{R}^{|C_{novel}|\times |C_{base}|}$ for the second $\emph{novel}$ split in PASCAL VOC. Figure \ref{fig:visualsim} shows the \emph{normalized} visual similarity $\mathbf{S}^{vis}(\mathbf{z}) \in \mathbb{R}^{|C_{base}|}$ for each proposal $\mathbf{z}$ (highlighted in \textcolor{blue}{blue}).

\vspace{-0.1in}

\section{Few-shot Performance on VOC's Base Classes}
\vspace{-0.05in}
\label{sec:baseclasses}
Due to lack of space, in the main paper, we focus on the detection/segmentation results obtained on the {\em novel} object classes; however, our model also learns to detect/segment {\em base} class objects as well.
We now illustrate that our proposed method improves performance on \emph{novel} classes as the value of $k$ is increased, without any significant performance drop on \emph{base} classes. The experimental setup and baselines are identical to the one described in Section \ref{exp:vocexp} of the main paper. Table \ref{tab:meta-rcnn-base} summarizes results on VOC \cite{everingham2010pascal} for the $1$-st novel split with $k$-shots, $k\in \{0, 3, 10\}$. 
%  We highlight that for the {\em novel} classes we use exactly the same amount of annotation as \cite{Kang_2019_ICCV} and \cite{Yan_2019_ICCV}, since weak image-level annotations required by our method can simply be derived from {\em base} object instance bounding box annotations.

% Our approach outperforms the related state-of-the-art methods on both \emph{novel} and \emph{base} classes. 
It is important to note that we are not using any additional annotations for the \emph{base} classes (when compared to \cite{Kang_2019_ICCV, Yan_2019_ICCV, wang2020frustratingly}). It can be seen that our model's performance on \emph{novel} classes steadily increases with increased $k$, while simultaneously retaining accuracy on \emph{base} classes.

Our slightly poorer performance on \emph{base} classes compared to \cite{wang2020frustratingly} can be attributed to the fact that \cite{wang2020frustratingly} use feature pyramid networks (FPN) \cite{lin2017feature} whereas we don't. According to \cite{lin2017feature}, FPN provides a performance improvement of about $3.8$ on AP$_{50}$ (See Table 3 in \cite{lin2017feature}, rows (a) and (c)). Our approach, despite not using FPNs, only performs $2.8$ points poorer on $k=0$. In addition, this gap reduces as the value of $k$ is increased ($1.3$ AP$_{50}$ on $k=3$; $0.7$ AP$_{50}$ on $k=10$). 
Also, compared to \cite{wang2020frustratingly}, UniT has a smaller drop in \emph{base} class performance as $k$ is increased. As an example, when $k$ is increased from $0$ to $5$, UniT has a performance drop of $0.2$ AP$_{50}$ on \emph{base} classes whereas \cite{wang2020frustratingly} has a larger drop of $1.7$ AP$_{50}$. 
This observation highlights the fact that our proposed approach is better at retaining \emph{base} class information compared to the existing state-of-the-art in \cite{wang2020frustratingly}. This improved retention can be mainly attributed to the structured decomposition of our detectors: weak detector + learned refinement. We believe that such a decomposition results in an inductive model bias that leads to convergence to an ultimately better solution. 
% Such decomposition may potentially be useful even in the traditional purely supervised setting
% \Sid{We believe that such a decomposition results in an inductive model bias that leads to convergence to an ultimately better solution. In other words, these results suggest, that such decomposition may potentially be useful even in the traditional purely supervised setting.}

% Table \ref{tab:meta-rcnn-base} highlights two key observations: i) Our performance on \emph{novel} classes steadily increases with increased $k$, and ii) Our performance on \emph{base} classes doesn't decrease significantly even when the value of $k$ is increased (decrease of $0.3$ AP$_{50}$ when the number of shots is 
% The significantly better performance on the \emph{base} classes can be mainly attributed to the structured decomposition of our detectors: weak detector + learned refinement. We believe that such a decomposition results in an inductive model bias that leads to convergence to an ultimately better solution. In other words, these results suggest, that such decomposition may potentially be useful even in the traditional purely supervised setting.  
% \raggedbottom
% possible that such a decomposition may be useful even in  traditional fully-supervised 
% is perhaps a useful way to approach the problem, leading to a better optimum.

% \input{tables/annotation-base}

\section {Additional Visualizations on MSCOCO Detection and Segmentation}
\label{sec:additionalviz}
\begin{figure*}[t]
\captionsetup[subfigure]{labelformat=empty}
\begin{subfigure}{.33\textwidth}
\centering
\includegraphics[width=\textwidth, height=3.5cm]{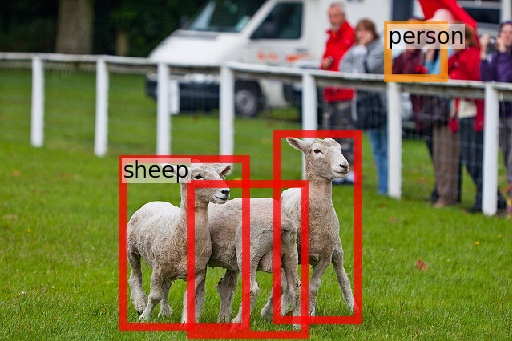}
\end{subfigure}%
\begin{subfigure}{.33\textwidth}
\centering
\includegraphics[width=\textwidth, height=3.5cm]{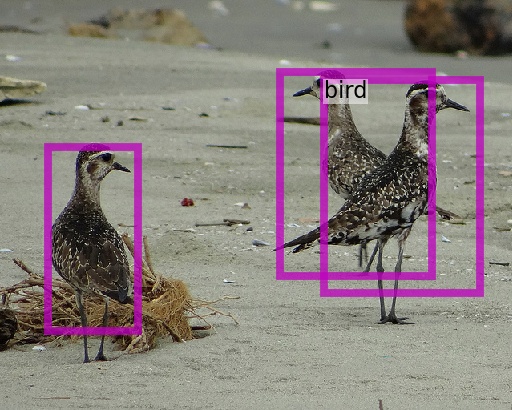}
\end{subfigure}%
\begin{subfigure}{.33\textwidth}
\centering
\includegraphics[width=\textwidth, height=3.5cm]{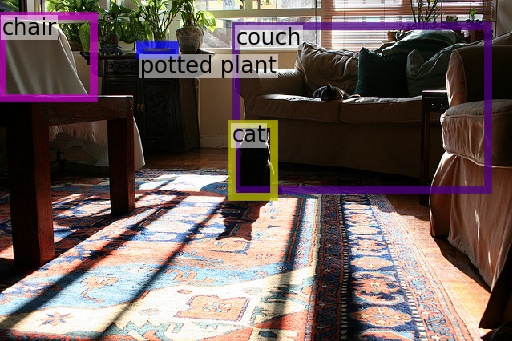}
\end{subfigure}
\begin{subfigure}{.33\textwidth}
\centering
\includegraphics[width=\textwidth, height=3.5cm]{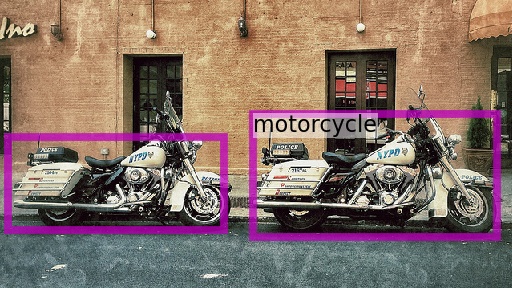}
\end{subfigure}%
\begin{subfigure}{.33\textwidth}
\centering
\includegraphics[width=\textwidth, height=3.5cm]{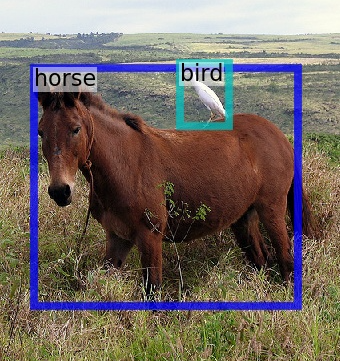}
\end{subfigure}%
\begin{subfigure}{.33\textwidth}
\centering
\includegraphics[width=\textwidth, height=3.5cm]{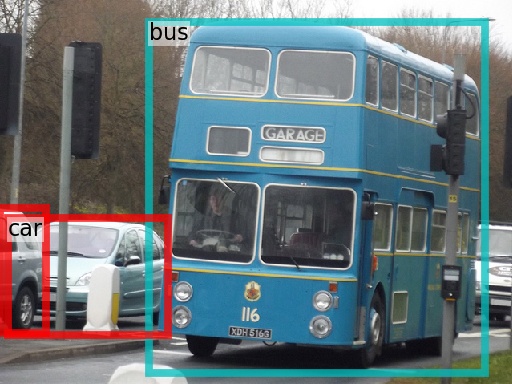}
\end{subfigure}
\begin{subfigure}{.33\textwidth}
\centering
\includegraphics[width=\textwidth, height=3.5cm]{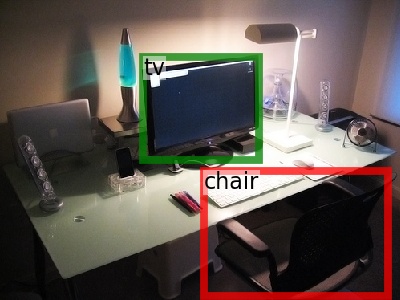}
\end{subfigure}%
\begin{subfigure}{.33\textwidth}
\centering
\includegraphics[width=\textwidth, height=3.5cm]{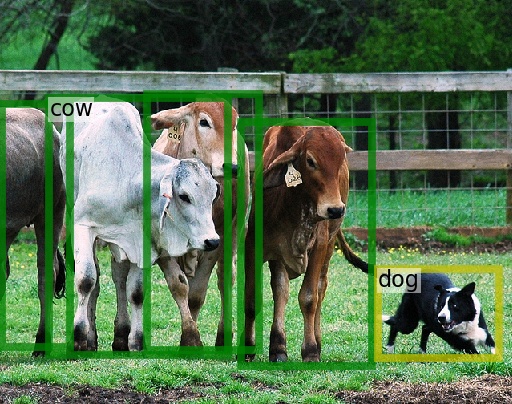}
\end{subfigure}%
\begin{subfigure}{.33\textwidth}
\centering
\includegraphics[width=\textwidth, height=3.5cm]{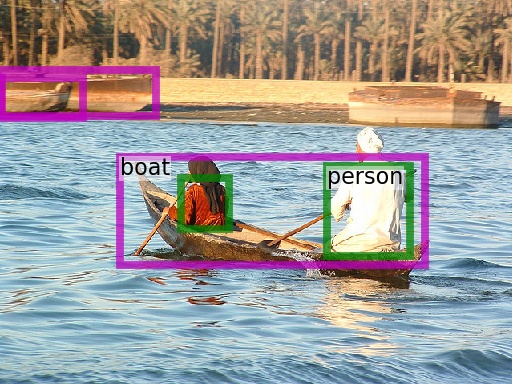}
\end{subfigure}
\begin{subfigure}{.33\textwidth}
\centering
\includegraphics[width=\textwidth, height=3.5cm]{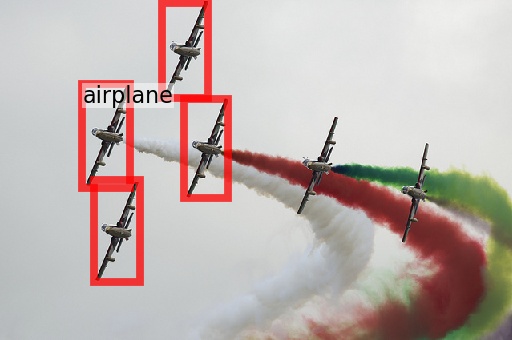}
\end{subfigure}%
\begin{subfigure}{.33\textwidth}
\centering
\includegraphics[width=\textwidth, height=3.5cm]{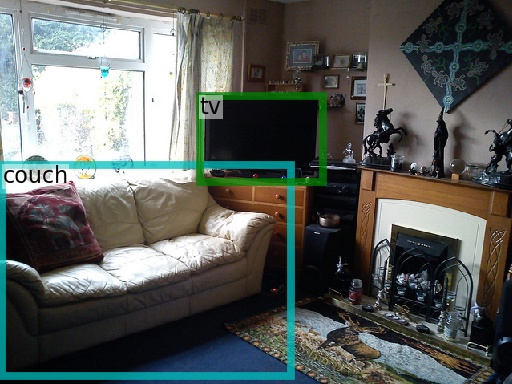}
\end{subfigure}%
\begin{subfigure}{.33\textwidth}
\centering
\includegraphics[width=\textwidth, height=3.5cm]{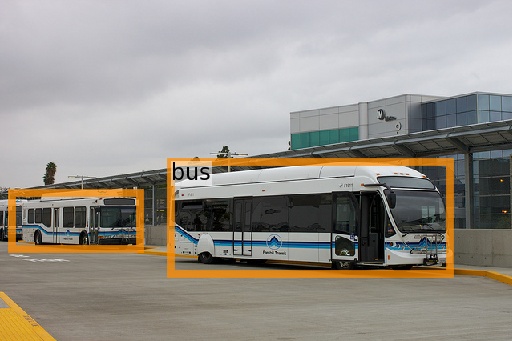}
\end{subfigure}
\begin{subfigure}{.33\textwidth}
\centering
\includegraphics[width=\textwidth, height=3.5cm]{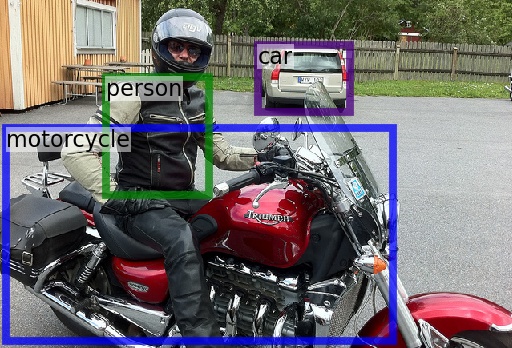}
\end{subfigure}%
\begin{subfigure}{.33\textwidth}
\centering
\includegraphics[width=\textwidth, height=3.5cm]{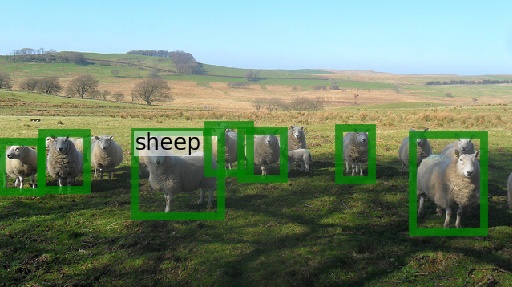}
\end{subfigure}%
\begin{subfigure}{.33\textwidth}
\centering
\includegraphics[width=\textwidth, height=3.5cm]{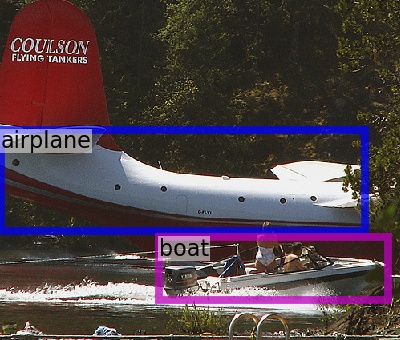}
\end{subfigure}
% \begin{subfigure}{.33\textwidth}
% \centering
% \includegraphics[width=\textwidth, height=3.5cm]{supp_figures/coco_det/26.jpg}
% \end{subfigure}
\vspace{-0.06in}
\caption{\textbf{Qualitative Visualizations.} Semi-supervised zero-shot ($k=0$) detection performance on \emph{novel} classes in MS-COCO (%best viewed with magnification; for each image, one 
color $=$ object category).}
\label{suppfig:cocoviz1}
\vspace{-0.2in}
\end{figure*}

\begin{figure*}[t]
\captionsetup[subfigure]{labelformat=empty}
\begin{subfigure}{.33\textwidth}
\centering
\includegraphics[width=\textwidth, height=3.5cm]{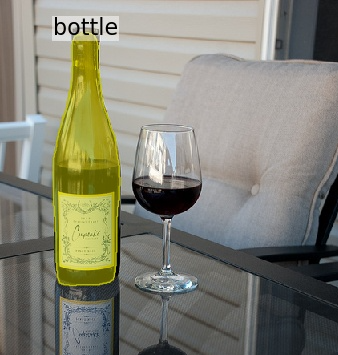}
\end{subfigure}%
\begin{subfigure}{.33\textwidth}
\centering
\includegraphics[width=\textwidth, height=3.5cm]{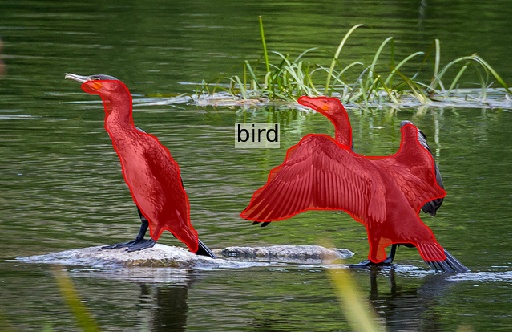}
\end{subfigure}%
\begin{subfigure}{.33\textwidth}
\centering
\includegraphics[width=\textwidth, height=3.5cm]{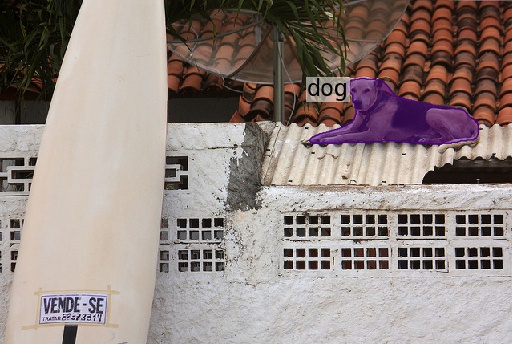}
\end{subfigure}
\begin{subfigure}{.33\textwidth}
\centering
\includegraphics[width=\textwidth, height=3.5cm]{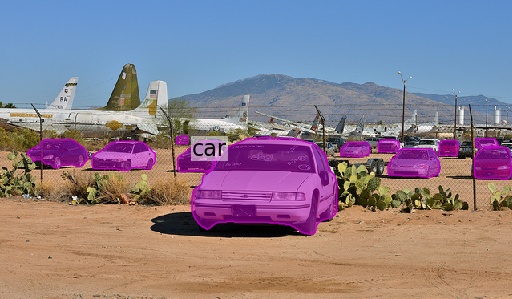}
\end{subfigure}%
\begin{subfigure}{.33\textwidth}
\centering
\includegraphics[width=\textwidth, height=3.5cm]{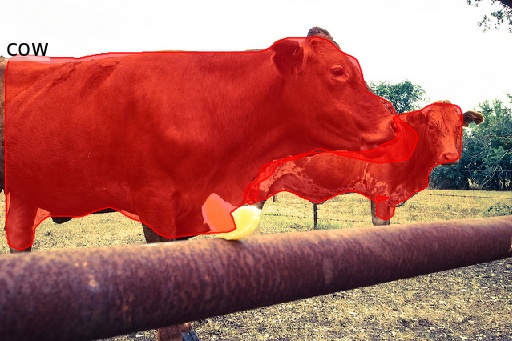}
\end{subfigure}%
\begin{subfigure}{.33\textwidth}
\centering
\includegraphics[width=\textwidth, height=3.5cm]{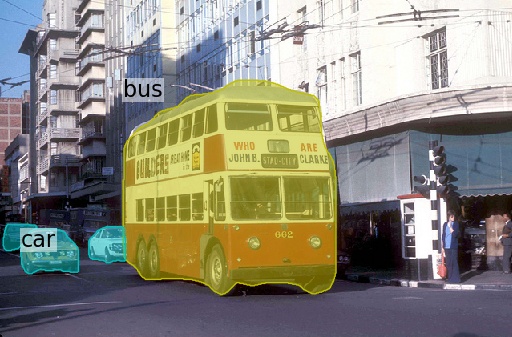}
\end{subfigure}
\begin{subfigure}{.33\textwidth}
\centering
\includegraphics[width=\textwidth, height=3.5cm]{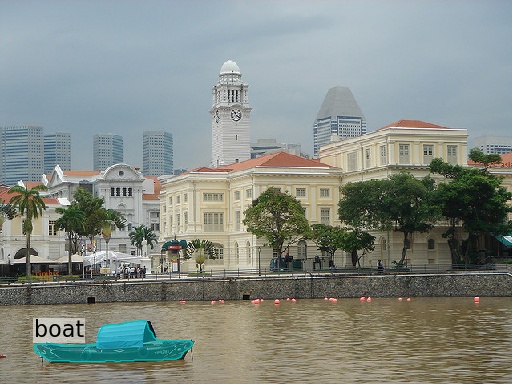}
\end{subfigure}%
\begin{subfigure}{.33\textwidth}
\centering
\includegraphics[width=\textwidth, height=3.5cm]{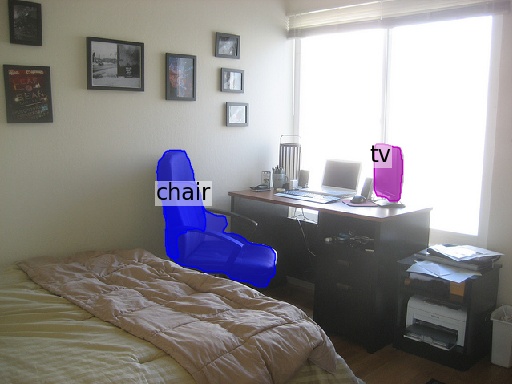}
\end{subfigure}%
\begin{subfigure}{.33\textwidth}
\centering
\includegraphics[width=\textwidth, height=3.5cm]{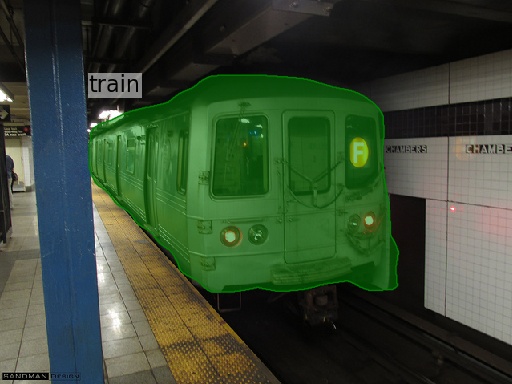}
\end{subfigure}
\begin{subfigure}{.33\textwidth}
\centering
\includegraphics[width=\textwidth, height=3.5cm]{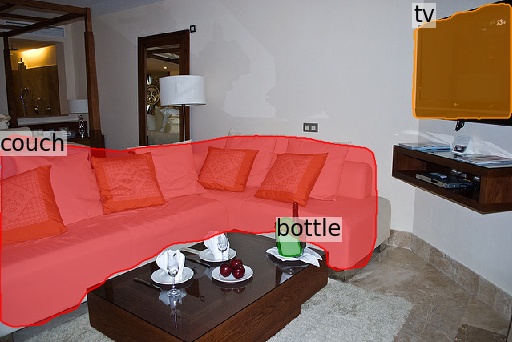}
\end{subfigure}%
\begin{subfigure}{.33\textwidth}
\centering
\includegraphics[width=\textwidth, height=3.5cm]{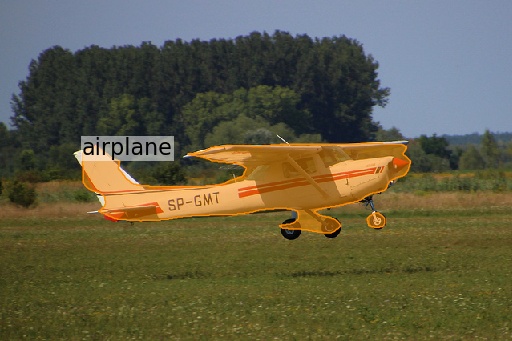}
\end{subfigure}%
\begin{subfigure}{.33\textwidth}
\centering
\includegraphics[width=\textwidth, height=3.5cm]{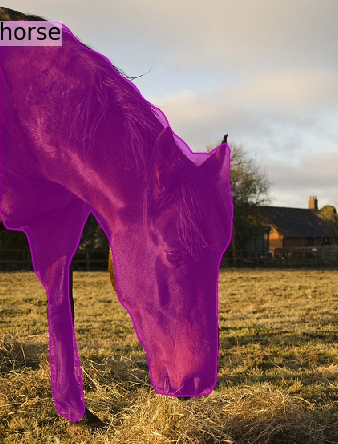}
\end{subfigure}
\begin{subfigure}{.33\textwidth}
\centering
\includegraphics[width=\textwidth, height=3.5cm]{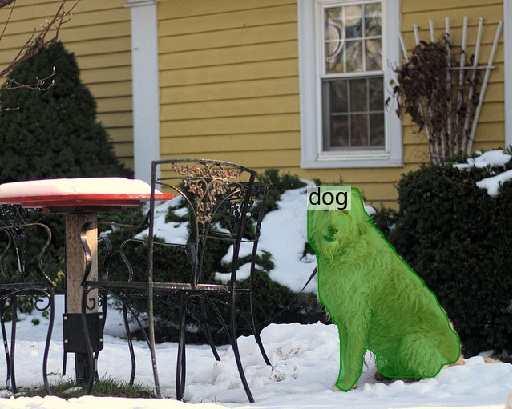}
\end{subfigure}%
\begin{subfigure}{.33\textwidth}
\centering
\includegraphics[width=\textwidth, height=3.5cm]{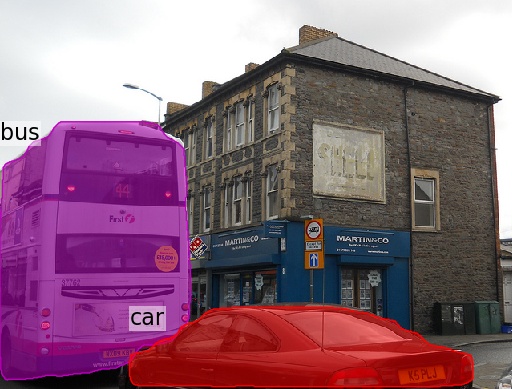}
\end{subfigure}%
\begin{subfigure}{.33\textwidth}
\centering
\includegraphics[width=\textwidth, height=3.5cm]{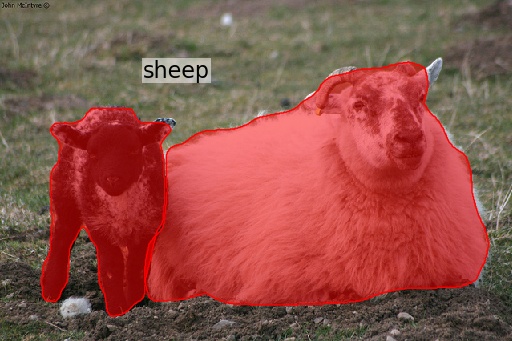}
\end{subfigure}
% \begin{subfigure}{.33\textwidth}
% \centering
% \includegraphics[width=\textwidth, height=3.5cm]{supp_figures/coco_seg/46.jpg}
% \end{subfigure}
\vspace{-0.06in}
\caption{\textbf{Qualitative Visualizations.} Semi-supervised zero-shot ($k=0$) instance segmentation performance on \emph{novel} classes in MS-COCO (%best viewed with magnification; for each image, one 
color $=$ object category).}
\label{suppfig:cocoviz2}
\end{figure*}

We show additional visualizations highlighting the performance of our approach on the MSCOCO \cite{lin2014microsoft} dataset. The experimental setup is identical to the ones described in Sections \ref{exp:vocexp} of the main paper. Figure \ref{suppfig:cocoviz1}
shows additional examples for the task of object detection, and Figure \ref{suppfig:cocoviz2} shows additional examples for the task of instance segmentation. Note that these visualizations are generated on \emph{novel} classes under the $k=0$ setup.

% \clearpage
% {
% % \raggedbottom
% \small
% \bibliographystyleAppendix{ieee_fullname}
% \bibliographyAppendix{egbib_appendix}
% \vspace{30em}
% \textcolor{white}{.}
% }
% {\small
% \bibliographystyle{ieee_fullname}
% \bibliography{egbib_appendix}
% }

% \input{tables/lwll_pres_table}
% \input{tables/ablation_table}

\end{document}